\documentclass[11pt]{article}

\usepackage[preprint]{acl}

\usepackage{times}
\usepackage{latexsym}

\usepackage[T1]{fontenc}

\usepackage[utf8]{inputenc}

\usepackage{microtype}

\usepackage{inconsolata}

\usepackage{graphicx}
\usepackage{booktabs}
\usepackage{multirow}
\usepackage{tabularx}

\usepackage{caption}
\usepackage{subcaption}
\usepackage{enumerate,enumitem}
\usepackage{amssymb}   
\usepackage{xcolor}    
\usepackage{amsmath}

\usepackage[table]{xcolor}

\newcommand{\medqaemo}{\ensuremath{\bullet}}
\newcommand{\mmluemo}{\ensuremath{\blacktriangle}}
\newcommand{\druidemo}{\ensuremath{\blacksquare}}

\newcommand\blfootnote[1]{%
  \begingroup
  \renewcommand\thefootnote{}\footnote{#1}%
  \addtocounter{footnote}{-1}%
  \endgroup
}

\title{A Reality Check of Inference-Time Context Augmentation Methods}

\def\authorsep{\hspace{0.3em}}

\author{Haeun Yu\textsuperscript{1*} \authorsep Geert Heyman\textsuperscript{2} \authorsep Pepa Atanasova\textsuperscript{1} \authorsep Isabelle Augenstein\textsuperscript{1} \medskip\\
\makebox[14em][c]{\llap{\textsuperscript{1}}University of Copenhagen}%
\makebox[14em][c]{\llap{\textsuperscript{2}}Nokia Bell Labs} \\
\makebox[14em][c]{Copenhagen, Denmark}%
\makebox[14em][c]{Antwerp, Belgium} \medskip\\
\texttt{hayu@di.ku.dk}}

\begin{document}
\maketitle
\begin{abstract}

Large language models (LLMs) often fail to utilise the context given in the input, resulting in hallucinations or a fallback to parametric knowledge. Context augmentation methods aim to improve model performance by modifying the input, from inserting filler tokens that add computational time to adding relevant content such as elaboration of a context during inference. However, these methods have been evaluated in isolation and only by their performance gains, leaving it unclear how they compare and whether they reliably improve performance. In this paper, we present an evaluation framework spanning five methods that differ in how informative the inserted content is and apply them at the sentence level across seven models spanning multiple sizes on three datasets. We find that accuracy gains are not always genuine improvements but undirected perturbations: the methods correct some predictions while breaking others, particularly when the inserted text allows only for more computational time. Our analysis reveals that methods that bring no additional content primarily change model predictions near the decision boundary. Our findings highlight the need to evaluate the reliability of the accuracy gains before adopting them in practice.

\end{abstract}

\section{Introduction}
\blfootnote{\textsuperscript{*} Work partially done during Haeun's internship at Nokia Bell Labs.}

Providing contextual knowledge during inference time allows large language models (LLMs) to better ground their answers in information, especially recent, domain-specific knowledge, without additional training. Despite the contextual knowledge being explicitly provided in the input, LLMs often fail to utilise it, resulting in hallucinations or defaulting to parametric knowledge \cite{wang2024resolving, xie2024adaptive, sun2025redeep}. To address this, a growing body of work proposes methods that insert additional content into the context at inference time, ranging from inserting filler tokens to augmenting the input with additional knowledge \cite{xu-etal-2024-reading, pfau2024lets, zhu-etal-2022-knowledge, chen-etal-2023-beyond}. We refer to these as \emph{context augmentation} methods, a subset of the broader family of context engineering methods.

\begin{figure}
    \centering
    \includegraphics[width=0.85\columnwidth]{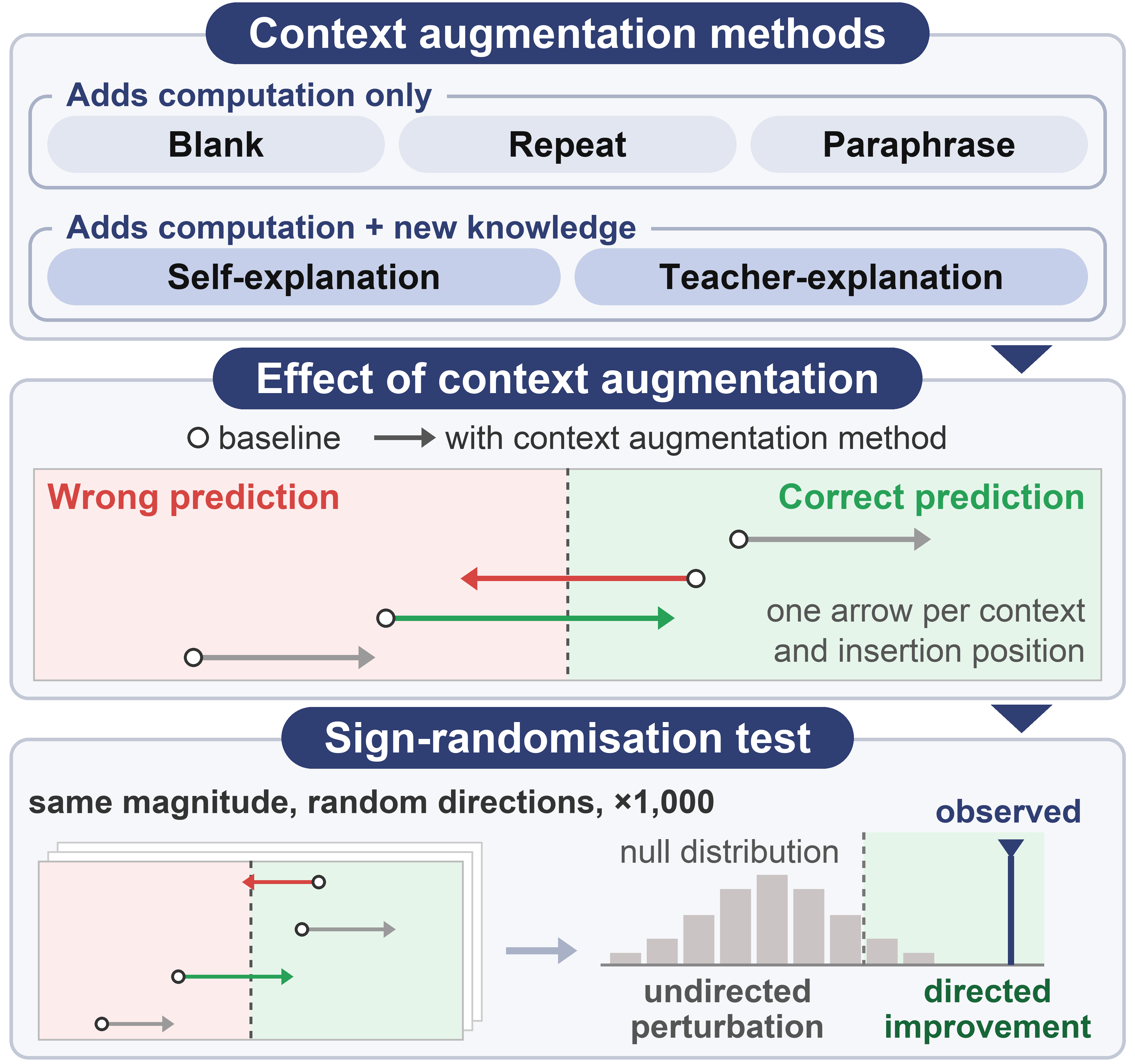}
    \caption{Our evaluation framework covers five methods, which can be split into two groups: one that only adds computation and the other that adds both computation and new content. While a method can change the model prediction, we further test whether the change is a directed improvement or undirected perturbation through the sign-randomisation test.}
    \label{fig:overview}
\end{figure}

However, these methods have been developed and evaluated in isolation and only by their performance gains. Performance gains alone do not explain whether the methods affect the model predictions in a reliable way. Furthermore, existing methods insert content at a limited, single, predetermined location, which differs across methods. To compare these methods systematically and understand their effects, we propose an evaluation framework that investigates the reliability of the accuracy gains across seven models and three datasets covering multiple-choice question answering (MCQA) and fact-checking.

Our evaluation framework covers five methods, each representing a family of context augmentation methods from prior work, as shown in Figure \ref{fig:overview}. We choose them to vary in how much information the inserted content carries, which allows us to compare \textit{how different amounts of information affect the model's predictions}. We further investigate how the intervention's position within the context affects its effectiveness. We therefore apply each method after every sentence of the context, one position at a time (Table~\ref{tab:example}). This provides a consistent set of intervention positions across methods and enables a \textit{systematic analysis of position-dependent effects}.

Our framework evaluates each method in two ways: \textit{by its accuracy, and by a sign-randomisation test} that assesses how reliably an accuracy gain reflects improvement. The test examines whether a gain stems from moving the model toward the correct answer given the context, or from perturbing its predictions without a consistent direction (\S\ref{sec:sign}). This distinction is necessary because such undirected perturbation can itself raise accuracy, reproducibly. We further analyse individual sentence-level interventions to understand which predictions each method changes (\S\ref{sec:margin}), at how many positions in a context (\S\ref{sec:multiplicity}), and whether successful positions can be identified in advance (\S\ref{sec:position}).

Our findings are:
\begin{itemize}[nosep]
    \item \textbf{Perturbations are undirected.} Accuracy gains from context augmentation methods do not always reflect a directed improvement (\S\ref{sec:results}). Particularly for methods with less information, they often arise from random changes, which we refer to as \textit{undirected perturbation}. The effect of a method also varies considerably across models and datasets.
    \item \textbf{Interventions act near the decision boundary.} Context augmentation methods with less information mainly change predictions that lie close to the decision boundary (\S\ref{sec:margin}). Whether they flip a prediction toward or away from the correct answer, they shift the prediction by similar amounts.
    \item \textbf{Intervention position is important.} Incorrect and correct predictions can be flipped from half of the positions in context (\S\ref{sec:multiplicity}). Successful positions, however, cannot be predicted in advance, neither from lexical characteristics of a sentence nor by using the last position, where content is commonly added (\S\ref{sec:position}). Identifying successful positions remains an open problem for future work.
\end{itemize}

\begin{table*}[t]
\centering
\small
\begin{tabular}{@{}p{2.8cm}p{\dimexpr\textwidth-2.8cm-2\tabcolsep\relax}@{}}
\toprule
\multicolumn{2}{@{}l}{\textbf{MMLU -- Professional Law \#753}} \\
\midrule
Context &
One day, while riding his motorcycle through a residential area, a man hit and injured a child who darted into his path.
The child's mother files suit against the man.
At trial, the mother is then called to testify as to her conversation with the man at the hospital where her son was rushed immediately following the accident.
\textbf{[Intervention text is inserted in this position.]}
According to the mother, the man told her, ``Don't worry, my insurance company will pay for all of your son's medical expenses.''
If offered into evidence, the court will most likely rule that the mother's statement is \\
\midrule
Options &
A.~admissible, on the issue of the defendant's negligence. \newline
B.~admissible, to show that the defendant is able to pay for the medical expenses. \newline
C.~inadmissible, because the testimony was highly prejudicial. \newline
\textbf{D.~admissible, only to prove ownership or control of the motorcycle.} \\
\midrule
\multicolumn{2}{@{}l}{\textbf{Intervention text}} \\
\addlinespace
Paraphrase &
At trial, the mother is then asked to testify about her conversation with the man at the hospital where her son was taken immediately after the accident. \\
\addlinespace
Self-explanation &
The mother's testimony is crucial in establishing the man's admission of liability and potential negligence in the accident. \\
\addlinespace
Teacher-explanation &
This means that during the court proceeding, the plaintiff (the mother) takes the witness stand and is asked to describe what the man said to her when they spoke at the hospital shortly after her son was taken there because of the accident. \\
\bottomrule
\end{tabular}
\caption{Example of intervention texts for an MMLU question. Each intervention is inserted at the marked position; the correct option is in bold.}
\label{tab:example}
\end{table*}

\section{Related Work}

Context augmentation methods modify the input to enhance model capability. For example, inserting filler tokens expands the computational space before answer generation, providing the model with additional computation to better process the input text \cite{pfau2024lets, goyal2024think, jang2025expanding}. Input repetition, on the other hand, enriches the model's internal representations by enabling bidirectional-like computation \cite{springer2025repetition}, and has been shown to improve performance on downstream language understanding tasks \cite{xu-etal-2024-reading, leviathan2025promptrepetition}.

Another line of work augments the context with additional knowledge rather than restructuring it. Retrieval-augmented generation (RAG) prepends retrieved passages to the input and has become a dominant paradigm for knowledge-intensive tasks \cite{RAG_original, zeng-etal-2025-towards}. Beyond retrieval from external corpora, several studies leverage the model itself or a better-performing model as a knowledge source, generating explanations or elaborations first and then incorporating them back into the input text \cite{liu-etal-2022-generated, chen-etal-2023-beyond, li-etal-2025-evaluating}. These methods also differ in where they intervene in the input: over the whole input \cite{xu-etal-2024-reading, leviathan2025promptrepetition}, between the context and the question \cite{liu-etal-2022-generated, li-etal-2025-evaluating}, or at the end of the input \cite{pfau2024lets, greenblatt2025filler}.

\citet{zhang2026illusionrobustness} examine how robust models are to task-irrelevant context by prepending it to benchmark questions, including semantically meaningless pseudo-words. They find that aggregate accuracy remains nearly unchanged while individual predictions flip in both directions, and call for per-example evaluation. Their study covers only the least informative kind of intervention; we compare it with methods that add increasing amounts of information and examine how their effects differ.

Overall, existing methods have been developed and evaluated in isolation and only by their performance gains. Because they intervene at different positions, practitioners have little guidance on when to apply which method and where. We address this gap with an evaluation framework that compares methods under the same conditions and analyses.

\section{Benchmarking Context Augmentation Methods}

Context augmentation methods intervene on the input text at inference time. We focus on methods that insert additional content into the input to improve context utilisation. These methods differ in how informative the inserted content is. Some add no new information and merely increase the amount of computation the model performs before answering, for example by inserting semantically irrelevant filler tokens \cite{pfau2024lets, greenblatt2025filler}. Others supply knowledge relevant to the question.

Existing methods have been proposed and evaluated in isolation, without systematic comparison. To compare them under the same conditions, we apply each method at the sentence level, standardising the intervention position across methods and allowing us to examine whether position matters. Our evaluation framework covers five methods, chosen to vary in how much information the inserted content carries, allowing us to compare how different amounts of information affect the model.

\subsection{Context Augmentation Methods}\label{sec:intervention}

For a context $c=[s_1, ..., s_n, ..., s_{N_c}]$ containing $N_c$ sentences, an intervention function $\mathcal{I}$ targets a sentence $s_n$ by appending an intervention $i_n$ after it. Formally, $\mathcal{I}(c, s_n)=[s_1, ...,s_n, i_n, ..., s_{N_c}]$. Each context is followed by a question, and the model outputs a token distribution over the vocabulary. Specifically, we compute the log probability of all possible answers and select the one with the highest log probability as the model answer. Our intervention approaches include five types: \textit{Blank}, \textit{Repeat}, \textit{Paraphrase}, \textit{Self-explanation}, and \textit{Teacher-explanation}, which are described below.

\paragraph{\textit{Blank}}

We append semantically irrelevant filler tokens after $s_n$, increasing the amount of computation the model performs over the context without adding any information. 

\paragraph{\textit{Repeat}}

A verbatim copy of $s_n$ is added. Beyond adding computation, repetition reinforces the original contextual information without introducing additional knowledge. This follows work showing that re-reading or repeating the input improves performance \cite{xu-etal-2024-reading, leviathan2025promptrepetition}.

\paragraph{\textit{Paraphrase}}

We insert a paraphrase of $s_n$ that preserves its meaning but uses different wording. Compared with \textit{Repeat}, the content is the same, but the surface form differs, allowing us to test whether restating information in different words affects the model beyond repetition.

\paragraph{\textit{Self-explanation}}

We insert an explanation of $s_n$ generated by the evaluated model itself. Since the relevant knowledge the model holds in its parameters is only implicit, generating it as text enables us to place it in the context, where the model can attend to it when answering \cite{yu2023generate, sun2023recitationaugmented, li-etal-2025-evaluating}. Unlike \textit{Paraphrase}, this adds information beyond the context, but only information the model already has.

\paragraph{\textit{Teacher-explanation}}
We insert an explanation of $s_n$ generated by a larger teacher model. It adds computation and information in the same way as \textit{Self-explanation}, but from a more capable model, and can therefore supply knowledge the evaluated model lacks.

\subsection{Evaluation Metrics}

Each intervened input contains a single intervention applied after one of the original sentences. A context $c$ with $N_c$ sentences thus yields $N_c$ separate intervened inputs per method, all paired with the same question. We look at individual intervened inputs to understand when and how each method affects the model prediction.

\subsubsection{Sentence-Averaged Accuracy}\label{sec:per_acc}

The sentence-averaged accuracy of a method is the fraction of its $N_c$ intervened inputs that are answered correctly, averaged over contexts:

\begin{equation}
\mathrm{Acc}_{\mathrm{sent}} = \frac{1}{|\mathcal{D}|} \sum_{c \in \mathcal{D}} \frac{1}{N_c} \sum_{n=1}^{N_c} y\big(\mathcal{I}(c, s_n)\big),
\end{equation}

where $\mathcal{D}$ is the set of contexts and $y(x) \in \{0, 1\}$ indicates whether the model answers the question correctly given input $x$. Without intervention, the accuracy is

\begin{equation}
\mathrm{Acc}_{\mathrm{orig}} = \frac{1}{|\mathcal{D}|} \sum_{c \in \mathcal{D}} y(c).
\end{equation}

The sentence-averaged accuracy decomposes as $\mathrm{Acc}_{\mathrm{sent}} = \mathrm{Acc}_{\mathrm{orig}} + \mathrm{Corr} - \mathrm{Brk}$, where $\mathrm{Corr}$ is the fraction of intervention positions at which the prediction flips from incorrect to correct, and $\mathrm{Brk}$ is the fraction at which it flips from correct to incorrect, both averaged over contexts in the same way as $\mathrm{Acc}_{\mathrm{sent}}$.

\subsubsection{Sign-Randomisation Test}\label{sec:sign}

The sign-randomisation test examines how reliably an accuracy gain reflects improvement. An accuracy gain can also arise from perturbing the model's predictions without a consistent direction. Because such a gain is deterministic and reproducible, neither standard significance tests nor variance across random seeds can separate it from a genuine improvement. We therefore test whether a method moves the model toward the correct answer more consistently than undirected perturbations of the same size.

For each input, we compute the margin $m$ between the log-probability of the correct option $o^{*}$ and that of the highest-scoring incorrect option:

\begin{equation}
m = \log p(o^{*}) - \max_{o \neq o^{*}} \log p(o),
\label{eq1}
\end{equation}

The margin is positive when the model answers correctly and negative when the answer is incorrect. For a context $c$, let $m_c$ denote the margin without intervention and $m_{c,n}$ the margin for $\mathcal{I}(c, s_n)$, i.e., with the intervention applied after sentence $s_n$. The change in margin caused by the intervention is $\Delta_{c,n} = m_{c,n} - m_c$. 

We set a null hypothesis that a method changes the margin without a consistent direction: it corrects some predictions and breaks others, so that the sign of each context's changes is arbitrary. We refer to such changes as \textit{undirected perturbations}. Rejecting the null hypothesis indicates that the method moves the model toward the correct answer more than expected by chance.

For each intervention position, we draw a random sign $\sigma_{c,n} \in \{-1, +1\}$ with equal probability and apply it to its change, giving sign-randomised margins $\tilde{m}_{c,n} = m_c + \sigma_{c,n}\,\Delta_{c,n}$. This preserves the magnitude of each change and removes only its direction. We recompute $\mathrm{Corr}$ and $\mathrm{Brk}$ on the sign-randomised margins and repeat this procedure $B = 1000$ times to obtain a random distribution for each.

A method rejects the null hypothesis when its observed $\mathrm{Corr}$ exceeds the 95\% interval of its random distribution while its observed $\mathrm{Brk}$ does not, i.e., it corrects incorrect predictions more often than undirected perturbations of the same size, without breaking correct predictions more often. An undirected perturbation that pushes predictions away from the current answer raises both $\mathrm{Corr}$ and $\mathrm{Brk}$, and therefore does not reject the null hypothesis.

\begin{table*}[t]
\centering
\resizebox{0.95\textwidth}{!}{%
\begin{tabular}{@{}lrlllll@{}}
\toprule
Model & \textbf{\textit{Original}} & \textbf{\textit{Blank}} & \textit{\textbf{Repeat}} & \textbf{\textit{Paraphrase}} & \textbf{\textit{Self-expl.}} & \textbf{\textit{Teacher-expl.}} \\
\midrule
OLMo-1b & 21.2 & \textbf{24.7}\,{\scriptsize $+$18.8/$-$15.3} & \textbf{24.4}\,{\scriptsize $+$19.9/$-$16.7} & \textbf{24.4}\,{\scriptsize $+$19.9/$-$16.7} & \textbf{24.9}\,{\scriptsize $+$18.4/$-$14.7} & \textbf{25.3}\,{\scriptsize $+$24.1/$-$20.0} \\
Llama-3b & 52.9 & 47.6\,{\scriptsize $+$5.4/$-$10.7} & 47.6\,{\scriptsize $+$5.9/$-$11.1} & 48.2\,{\scriptsize $+$6.2/$-$11.0} & 49.6\,{\scriptsize $+$10.1/$-$13.4} & \cellcolor{teal!35}\textbf{54.9}\,{\scriptsize $+$12.4/$-$10.4} \\
Gemma-4b & 45.7 & \cellcolor{teal!35}\textbf{46.6}\,{\scriptsize $+$2.1/$-$1.1} & \textbf{46.8}\,{\scriptsize $+$2.8/$-$1.7} & \textbf{46.9}\,{\scriptsize $+$3.0/$-$1.8} & \textbf{48.1}\,{\scriptsize $+$8.7/$-$6.2} & \textbf{55.0}\,{\scriptsize $+$14.1/$-$4.7} \\
OLMo-7b & 40.6 & \textbf{40.8}\,{\scriptsize $+$3.2/$-$3.0} & \textbf{40.8}\,{\scriptsize $+$3.5/$-$3.2} & \textbf{40.7}\,{\scriptsize $+$3.5/$-$3.3} & \textbf{43.0}\,{\scriptsize $+$9.4/$-$7.0} & \textbf{47.6}\,{\scriptsize $+$15.5/$-$8.5} \\
Llama-8b & 53.6 & 46.3\,{\scriptsize $+$5.2/$-$12.5} & 43.2\,{\scriptsize $+$3.9/$-$14.4} & 43.6\,{\scriptsize $+$4.2/$-$14.2} & 49.3\,{\scriptsize $+$9.4/$-$13.7} & 52.2\,{\scriptsize $+$10.7/$-$12.1} \\
Gemma-27b & 67.0 & \textbf{67.7}\,{\scriptsize $+$1.3/$-$0.7} & \cellcolor{teal!35}\textbf{67.7}\,{\scriptsize $+$2.0/$-$1.3} & \cellcolor{teal!35}\textbf{67.5}\,{\scriptsize $+$1.9/$-$1.4} & \textbf{69.5}\,{\scriptsize $+$5.9/$-$3.4} & \cellcolor{teal!35}\textbf{73.9}\,{\scriptsize $+$9.7/$-$2.8} \\
Qwen-72b & 73.4 & \textbf{73.4}\,{\scriptsize $+$1.4/$-$1.4} & 72.7\,{\scriptsize $+$1.2/$-$1.9} & 72.7\,{\scriptsize $+$1.4/$-$2.0} & \cellcolor{teal!35}\textbf{73.7}\,{\scriptsize $+$3.4/$-$3.1} & \cellcolor{teal!35}\textbf{76.8}\,{\scriptsize $+$6.3/$-$2.9} \\
\bottomrule
\end{tabular}%
}
\caption{Sentence-averaged accuracy (\%) of each method on MedQA. \textit{Original} reports an accuracy without intervention. \textbf{Bold} numbers mark accuracies higher than the \textit{Original}. Small numbers with $+$ are the fraction of intervention positions at which the prediction flips from incorrect to correct ($\mathrm{Corr}$), and those with $-$ are the fraction of the positions at which the prediction flips from correct to incorrect ($\mathrm{Brk}$). We colour cells where the method corrects wrong predictions more often than expected under undirected perturbation, according to the sign-randomisation test (\S\ref{sec:sign_result}).}
\label{tab:acc_medqa}
\end{table*}

\section{Experimental Setup}

We apply our evaluation framework for context augmentation methods to seven open-weight models across three different datasets. 

\subsection{Datasets}

As our evaluation framework examines where in the context an intervention should be applied, we select datasets where gold answers are not explicit in contexts. If a single sentence contained the answer, it would trivially be the most effective position to intervene. To this end, we study our framework with MedQA \cite{medqa}, an MCQA dataset for the medical domain, MMLU \cite{hendrycks2021measuring}, an MCQA dataset for world knowledge, and DRUID \cite{hagstrom-etal-2025-reality}, a fact-checking dataset. For those datasets, the context provides background that is necessary to understand the question but does not reveal the answer.

Most MMLU questions consist of a single short sentence without an accompanying context, which leaves no sentence positions at which to intervene. We therefore restrict MMLU to five subjects whose questions include a multi-sentence passage before the question itself. For DRUID, we exclude claims with insufficient evidence, since a method can fail because of context quality, not the method itself. We provide details for each dataset and a subject selection for MMLU in Appendix \ref{appendix:dataset}.

\subsection{Implementation of Interventions}

For the \textit{Blank} intervention, we set the intervention text $i_n$ to a sequence of whitespace characters matching the character length of $s_n$. For \textit{Repeat}, we set $ i_n = s_n$, repeating the sentence. We obtain paraphrased sentences for \textit{Paraphrase} by strictly prompting GPT-5.4-Mini\footnote{\url{https://developers.openai.com/api/docs/models/gpt-5.4-mini}}, without introducing additional information. For \textit{Self-explanation}, $i_n$ is an elaboration of $s_n$ that is generated by the same model being evaluated. For generating an intervention text for \textit{Teacher-explanation}, we query the teacher model to generate an elaboration of $s_n$. We use the same GPT-5.4-Mini model as above. To prevent explanations from offering a shortcut to the answer (e.g. simply stating the correct answer), the generating model sees only the context, without the question or answer options. We use the same input prompt for querying self-explanation. We provide the prompts used to generate intervention texts in Appendix \ref{appx:implementation}. Table \ref{tab:example} shows an example of intervention texts with original context and question. For all methods, we utilise the existing evaluation scheme, which standardises evaluation across different datasets \cite{eval-harness}.

\subsection{Models}

We evaluate seven open-weight LLMs from four model families for reproducibility: OLMo-2-0425-1B-Instruct (OLMo-1b, \cite{olmo20252olmo2furious}), Olmo-3-7B-Instruct (OLMo-7b, \cite{olmo2025olmo3}), Llama-3.2-3B-Instruct (Llama-3b, \cite{grattafiori2024llama3herdmodels}), Meta-Llama-3.1-8B-Instruct (Llama-8b, \cite{grattafiori2024llama3herdmodels}), Gemma-3-4b-it (Gemma-4b, \cite{Gemmateam2025Gemma3technicalreport}), Gemma-3-27b-it (Gemma-27b, \cite{Gemmateam2025Gemma3technicalreport}), and Qwen2.5-72B-Instruct-AWQ (Qwen-72b, \cite{qwen2.5}).

\begin{table*}[]
\centering
\resizebox{\textwidth}{!}{%
\begin{tabular}{@{}l|lllllll@{}}
\toprule
\textbf{} & \textbf{OLMo-1b} & \textbf{Llama-3b} & \textbf{Gemma-4b} & \textbf{OLMo-7b} & \textbf{Llama-8b} & \textbf{Gemma-27b} & \textbf{Qwen-72b} \\ \midrule
\textit{\textbf{Blank}} & - & \druidemo{} & \medqaemo{}, \druidemo{} & - & \mmluemo{} & - & - \\
\textit{\textbf{Repeat}} & \druidemo{} & \druidemo{} & - & - & - & \medqaemo{} & \druidemo{} \\
\textit{\textbf{Paraphrase}} & - & - & - & - & - & \medqaemo{} & \druidemo{} \\
\textit{\textbf{Self-explanation}} & \druidemo{} & - & - & \mmluemo{} & \mmluemo{}, \druidemo{} & - & \medqaemo{}, \druidemo{} \\
\textit{\textbf{Teacher-explanation}} & - & \medqaemo{}, \mmluemo{}, \druidemo{} & - & - & \mmluemo{}, \druidemo{} & \medqaemo{} & \medqaemo{}, \druidemo{} \\ \bottomrule
\end{tabular}%
}
\caption{Sign-randomisation test results across all context augmentation methods and models. We mark the datasets on which the observed corrected proportion exceeds the 95\% interval of the null distribution while the broken proportion does not. \medqaemo{} denotes the MedQA dataset, \mmluemo{} MMLU, \druidemo{} DRUID. - indicates that no dataset passes the test.}
\label{tab:sign_random}
\end{table*}

\section{Results}\label{sec:results}

\subsection{Sentence-Averaged Accuracy Results}

Table \ref{tab:acc_medqa} reports sentence-averaged accuracy of the context augmentation methods (\S\ref{sec:intervention}) on MedQA. Tables \ref{tab:acc_mmlu} and \ref{tab:acc_druid} (Appendix \ref{appx:results}) report sentence-averaged accuracies on MMLU and DRUID. 

Methods that add new information yield the largest gains on MedQA. \textit{Teacher-explanation} improves sentence-averaged accuracy over \textit{Original} for six of seven models, by 2.0 to 9.3 points, and \textit{Self-explanation} by 0.3 to 3.7 points for five models. However, the uncoloured cells show that these gains are not necessarily directed. The largest gains, \textit{Teacher-explanation} with Gemma-4b (+9.3) and OLMo-7b (+7.0), do not reject the null hypothesis. Although $\mathrm{Corr}$ exceeds $\mathrm{Brk}$, because $\mathrm{Brk}$ significantly exceeds what undirected perturbations of the same size would produce, the gain does not reflect a consistent move toward the correct answer.

Methods that add no content (\textit{Blank}, \textit{Repeat}, and \textit{Paraphrase}) barely change sentence-averaged accuracy for most models: $\mathrm{Corr}$ and $\mathrm{Brk}$ are of similar size and cancel each other out (e.g., +3.2/$-$3.0 for \textit{Blank} with OLMo-7b). Accuracy alone cannot distinguish such cancelling from a method having no effect: for example, \textit{Teacher-explanation} with OLMo-7b on MMLU changes sentence-averaged accuracy by only 0.1 points, while changing the prediction for 19.1\% of intervention positions.

Overall, applying context augmentation at the sentence level shows that accuracy gains are inconsistent across models and datasets, and that many of them arise from undirected perturbation rather than from moving the model toward the correct answer. This highlights the need to complement accuracy-based evaluation with tests of direction, such as our sign-randomisation test.

\subsection{Sign-Randomisation Test Results}\label{sec:sign_result}

Table \ref{tab:sign_random} marks the datasets on which a method corrects wrong predictions more often than expected under undirected perturbation. Methods that add new information reject the null hypothesis more often, indicating they move the models toward the correct answer more than expected by chance. \textit{Teacher-explanation} does so in 8 of 21 model–dataset pairs and \textit{Self-explanation} in 6, compared with 4 for \textit{Repeat} and 4 for \textit{Blank}. Results also vary across models: while undirected perturbation prevails for OLMo models, Qwen-72b and Llama models reject the null hypothesis most often.

This is consistent with prior observations that context interventions can change predictions in both directions \citep{li-etal-2025-evaluating, zhang2026illusionrobustness}. Appendix \ref{appx:results} reports random distributions for both $\mathrm{Corr}$ and $\mathrm{Brk}$, including the difference between observed and null accuracy.

\begin{figure*}
     \centering
     \begin{subfigure}[b]{0.47\textwidth}
         \centering
         \includegraphics[width=\columnwidth]{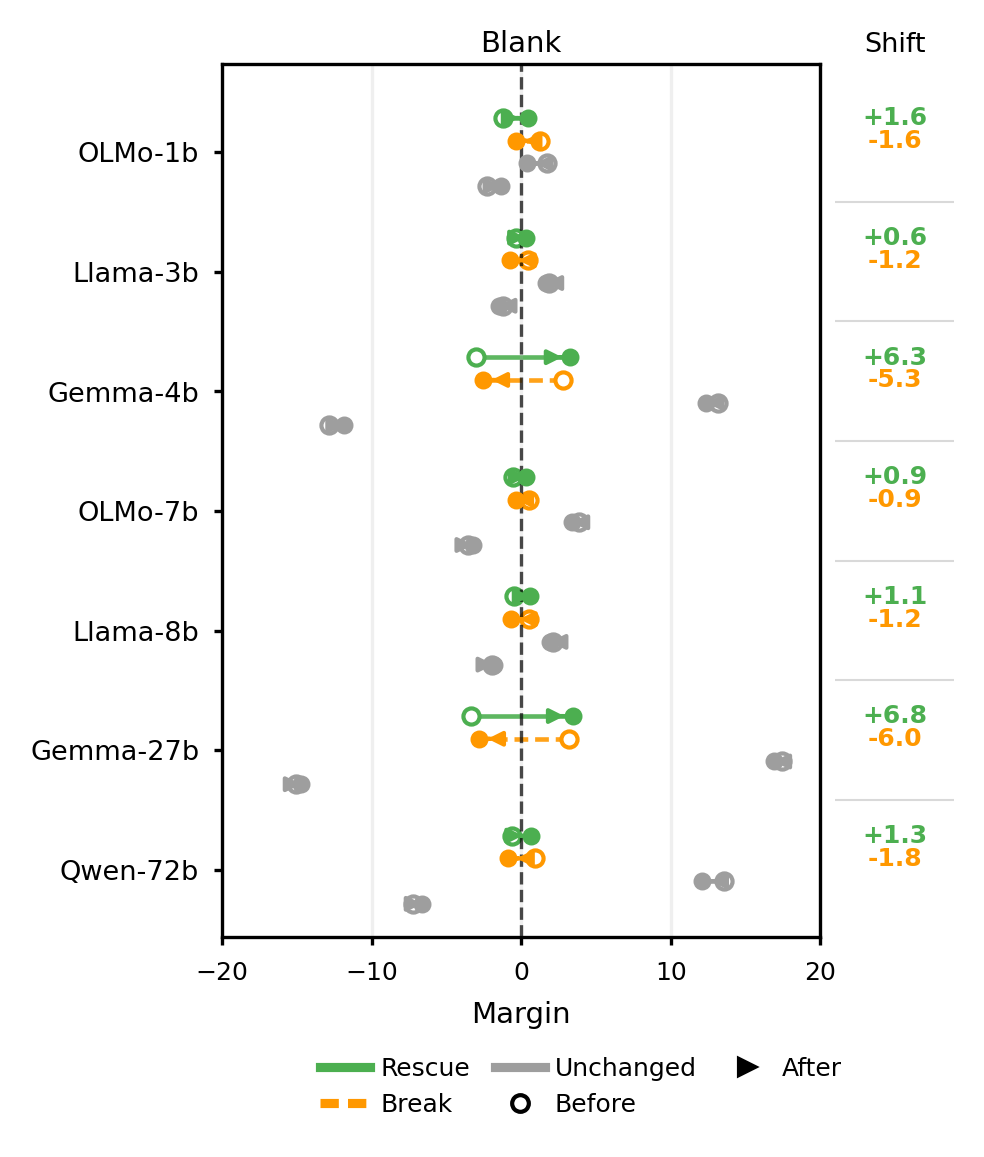}
     \end{subfigure}
     \hfill
     \begin{subfigure}[b]{0.47\textwidth}
         \centering
         \includegraphics[width=\columnwidth]{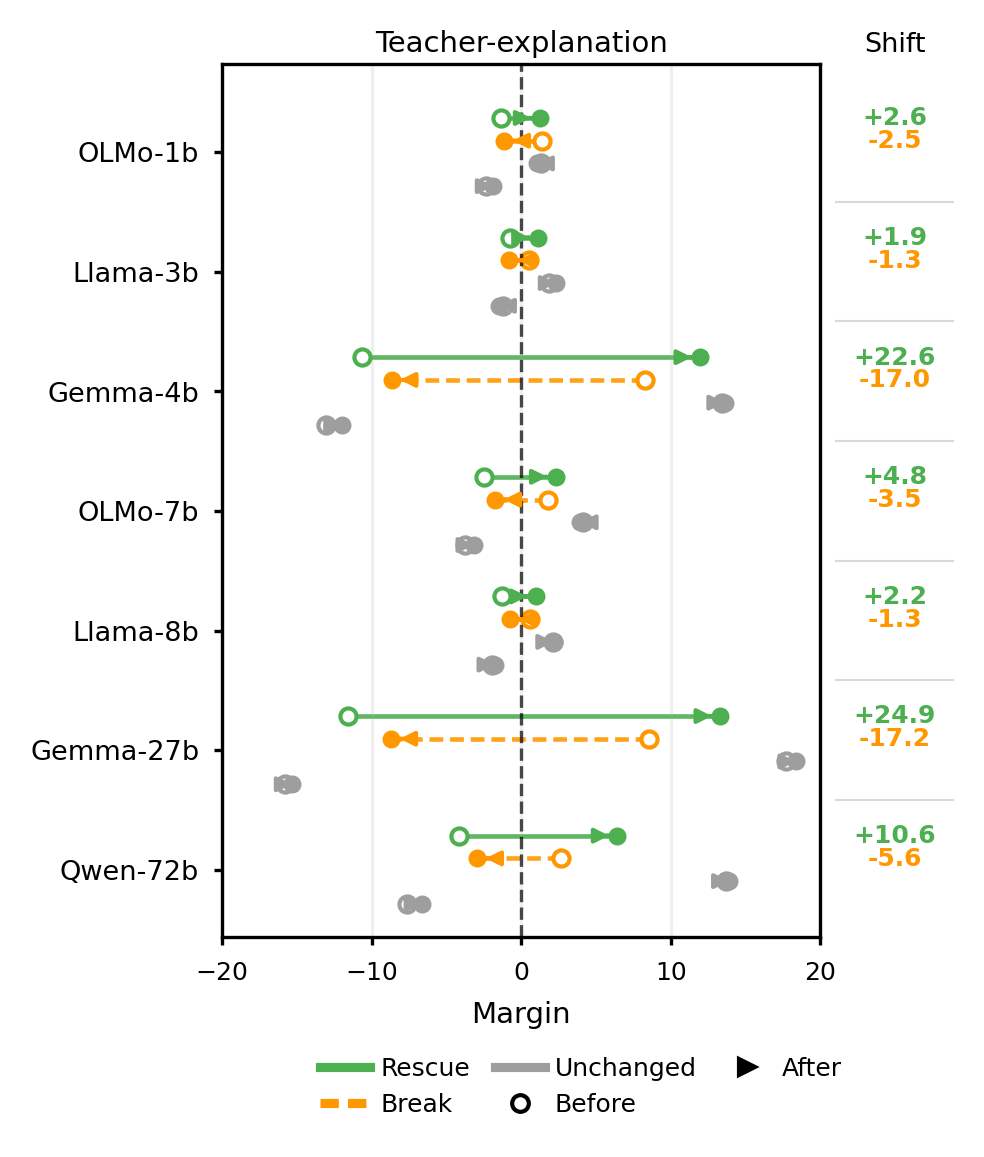}
     \end{subfigure}
        \caption{We show how margin changes before (\textit{Original}, hollow circles) and after (filled circles) intervention, averaged within each category on MedQA across seven models with \textit{Blank} and \textit{Teacher-explanation}. For each model, the first row and the second row are \texttt{Rescue} and \texttt{Break} and the last two rows are \texttt{Unchanged}.}
        \label{fig:margin_plots}
\end{figure*}

\section{Analysis}\label{sec:analysis}

We analyse individual interventions on each sentence to understand for which instances (\S\ref{sec:margin}), how consistently (\S\ref{sec:multiplicity}), and at which positions in the context (\S\ref{sec:position}) each method changes the model's prediction.

\subsection{Which Instances are Affected?}\label{sec:margin}

We examine how the margin (Equation \ref{eq1}) changes from \textit{Original} when a context augmentation method is applied at a single sentence position to understand which instances the methods affect. We classify each intervention as \texttt{Rescue}, \texttt{Break}, or \texttt{Unchanged}. \texttt{Rescue} is when an intervention turns an incorrect prediction into a correct one. \texttt{Break} is when an intervention turns a correct prediction into an incorrect one. \texttt{Unchanged} is when an intervention does not change the model's prediction, including both correct and incorrect ones.

Figure \ref{fig:margin_plots} presents results for \textit{Blank} and \textit{Teacher-explanation} on MedQA. Results for other methods and datasets are in Appendix \ref{appx:margin}. For most models on MedQA, \texttt{Rescue} and \texttt{Break} involve instances near the decision boundary (dashed line, margin = 0). We also note that both categories change the margin by similar amounts in opposite directions. This trend holds across all datasets. This implies that methods change predictions mainly where the model is already uncertain. In contrast, \texttt{Unchanged} groups lie far from the boundary, and their margins shift only slightly on average.

Comparing methods shows that margin-size changes depend on the method. Knowledge-oriented methods shift the margin more strongly than computation-oriented methods, in both directions.

\subsection{How Many Positions are Affected?}\label{sec:multiplicity}

Sentence-averaged accuracy shows how often a single intervention leads to the correct answer, but not how many opportunities exist within a context. This determines how much a method's advantage depends on choosing the position. If most positions succeed, any choice is likely to work, whereas if only one works, identification of the position becomes important.

We measure multiplicity of methods along two axes, \texttt{Rescue} and \texttt{Break}. In Table \ref{tab:multiplicity}, the \texttt{Rescue} column shows the fraction of positions at which an intervention turns an incorrect prediction into a correct one, among contexts rescued at least once. The \texttt{Break} column shows the fraction of positions at which it turns a correct prediction into an incorrect one, among contexts broken at least once. We average across all methods and models; the detailed results are presented in Appendix \ref{appx:multi}.

The \texttt{Rescue} fraction is similar to the \texttt{Break} fraction across datasets and models. The high standard deviation on MedQA is mainly due to OLMo-1b (Table~\ref{tab:multi_medqa}). Once a context can be affected, the methods change the prediction at about half of its sentence positions, regardless of whether the original prediction was correct. The chosen position, therefore, determines whether a method helps or harms, so knowing which positions are likely to succeed is important to maximise its benefit. This motivates the question of whether successful positions can be identified in advance (\S\ref{sec:position}).

\begin{table}[]
\centering
\resizebox{0.7\columnwidth}{!}{%
\begin{tabular}{@{}ccc@{}}
\toprule
Dataset & \texttt{Rescue} rate & \texttt{Break} rate  \\ \midrule
MedQA & 0.553 $\pm$ 0.18 & 0.569 $\pm$ 0.23 \\
MMLU & 0.422 $\pm$ 0.09 & 0.422 $\pm$ 0.09 \\
DRUID & 0.459 $\pm$ 0.11 & 0.411 $\pm$ 0.11 \\ \bottomrule
\end{tabular}%
}
\caption{Multiplicity analysis. \texttt{Rescue} rate: fraction of sentence positions at which an intervention rescues the prediction, among contexts rescued at least once. \texttt{Break} rate: the same for breaking, among contexts broken at least once. Values are means $\pm$ standard deviations over all methods and models.}
\label{tab:multiplicity}
\end{table}

\subsection{Can Positions be Identified?}\label{sec:position}

We examine if a successful position can be identified from two perspectives: lexical characteristics of the target sentence and a fixed position.

\paragraph{Lexical characteristics.} We first look at lexical characteristics of a sentence, which are cheap to obtain, as they require no model access. We follow the taxonomy of context characteristics of \citet{hagstrom-etal-2025-reality}, apply it at the sentence level, and add coreference characteristics, as context augmentation methods may help the model resolve references across sentences. Table \ref{tab:lexical_characteristics} lists ten characteristics used in the analysis. We extract coreference chains with the coreferee library.\footnote{\url{https://github.com/richardpaulhudson/coreferee}} For each characteristic, we compute AUROC across all interventions: the probability that a sentence where the intervention leads to \texttt{Rescue} (or \texttt{Break}) has a higher value than another sentence across all contexts where it does not. All AUROCs are near chance (0.48--0.54; Table \ref{tab:lexical_auroc}), indicating that lexical characteristics alone cannot identify successful intervention positions. We describe results in detail in Appendix \ref{appx:lexical}.

\begin{figure}
    \centering
         \includegraphics[width=0.9\columnwidth]{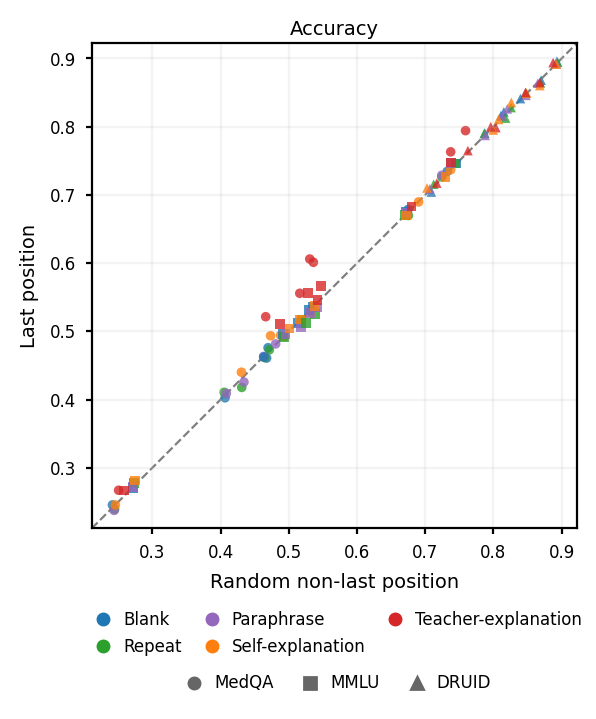}
         \caption{Accuracy when intervening after the last sentence (y-axis) versus after a randomly selected non-last sentence (x-axis; mean over three random draws). Points close to the dashed diagonal indicate that the last position is no better than other positions.}
         \label{fig:last_rest}
\end{figure}

\paragraph{Fixed position.} We compare accuracy when intervening at the last sentence, directly before the question, with accuracy at a randomly selected position among the rest of the positions except the last position (Random non-last position). We randomly select a position three times and average the accuracy. For every method, model, and dataset, performance at the last position is mostly on par with the other positions, with all points lying on the diagonal except for some \textit{Teacher-explanation} results (Figure \ref{fig:last_rest}). Although intervening at the end of the input is common practice \cite{liu-etal-2022-generated}, our results suggest that other positions perform as well on average.

Together with the multiplicity analysis (\S\ref{sec:multiplicity}), these results show that successful positions are plentiful but cannot be identified in advance, neither from a sentence's content nor from its location. 

\section{Discussion \& Conclusion}

Prior work reports that context augmentation methods, from inserting filler tokens to adding knowledge, improve performance \cite{liu-etal-2022-generated, pfau2024lets, li-etal-2025-evaluating}. Our sentence-averaged accuracy results (Tables \ref{tab:acc_medqa}, \ref{tab:acc_mmlu}, and \ref{tab:acc_druid}) also show accuracy gains, though not in all settings. This is consistent with \citet{li-etal-2025-evaluating}, who find that the benefit of self-generated documents, a variant of \textit{Self-explanation}, depends on the document format and dataset. Our results extend this to other methods and also show that a method's effect is inconsistent across datasets and depends on the model.

However, accuracy gains alone cannot distinguish a method that guides the model toward the correct answer from one that merely perturbs its prediction. \citet{zhang2026illusionrobustness} make a related point, showing that predictions resulting from task-irrelevant interventions flip in both directions even when aggregate accuracy is stable. Our sign-randomisation test shows that such undirected perturbation is not limited to methods that only add computation, but also occurs for methods that add content. Furthermore, our multiplicity analysis offers a finer-grained view of the findings of \citet{zhang2026illusionrobustness}: once a prediction can be changed, it is rescued or broken at similar fractions of sentence positions (\S\ref{sec:multiplicity}).

Context augmentation methods should therefore not be applied blindly. Our framework allows researchers and practitioners to verify whether a method's gains are directed and how often it breaks correct predictions. Our results favour methods that add new information, validated for each model family and dataset. Since the chosen position determines whether an intervention corrects or breaks a prediction, identifying successful positions in advance, and thereby maximising the benefits of context augmentation methods, remains an open direction for future work.

\section*{Limitations}

Our datasets are multiple-choice and binary classification tasks, and we score answer options by log-likelihood rather than generating free-form text. Our findings may therefore not transfer to open-ended generation. We also apply one intervention at a time, after a single sentence, to isolate the effect of each position. Prior work typically applies these methods once to the whole context, and gains may differ in that setting or when several interventions are combined. Our evaluation framework, however, can be extended to any intervention that changes the model’s margin.

We evaluate seven instruction-tuned open-weight models from 1B to 72B parameters. To avoid confounding the interventions with model-generated reasoning, we do not include reasoning models, whose own chain of thought may interact with the inserted text. All datasets are in English.

We used AI assistants for grammar checking and stylistic suggestions, and for help with writing analysis and plotting code. No generated text was used without revision, and all code and results were reviewed and checked by the authors.

\section*{Acknowledgements}

$\begin{array}{l}\includegraphics[width=1cm]{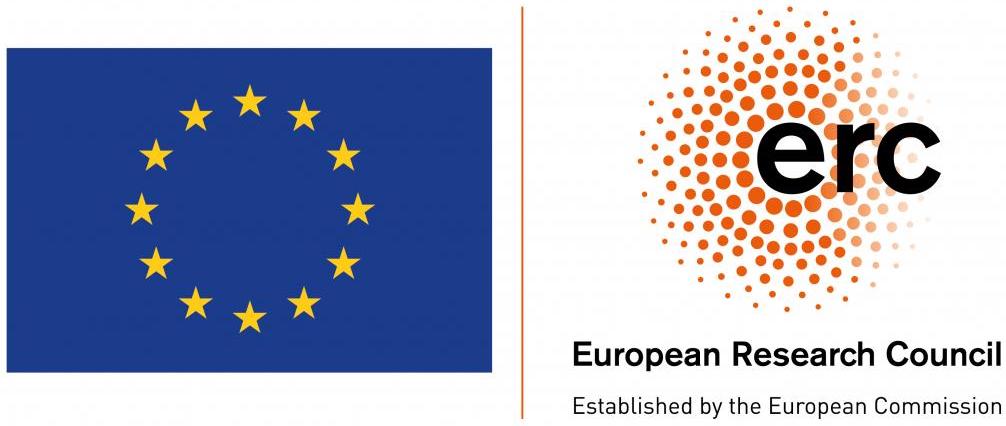} \end{array}$ 
This research was funded by the European Union (ERC, ExplainYourself, 101077481) and supported by the Pioneer Centre for AI, DNRF grant number P1.

Views and opinions expressed are however those of the author(s) only and do not necessarily reflect those of the European Union or the European Research Council. Neither the European Union nor the granting authority can be held responsible for them.

\bibliography{custom}

\appendix

\section{Experimental Details}

\subsection{Dataset Details}\label{appendix:dataset}

MedQA is a multiple-choice Question Answering (MCQA) dataset in which a model is required to select the correct option given a patient description as context and a question. MMLU is an MCQA dataset, which examines models' world knowledge and problem-solving ability. Finally, DRUID is a fact-checking dataset containing claims and evidence as contexts, retrieved from the real world. We only take instances with support and refute labels, making it a binary classification.

For MMLU, we select the five subjects whose questions exceed 500 characters on average: high school European, US and world history, professional law, and professional medicine (654–1,391 characters). All remaining subjects average below 330 characters, so the threshold clearly separates the two groups.

\begin{table}[]
\centering
\resizebox{\columnwidth}{!}{%
\begin{tabular}{@{}crrr@{}}
\toprule
Dataset & Number of Questions & Mean $\pm$ SD     & Total sentences \\ \midrule
MedQA   & 1,270               & 7.09 $\pm$ 3.28 & 8,998           \\
MMLU    & 2,310               & 6.52 $\pm$ 3.84 & 15,063          \\
DRUID   & 2,669               & 5.72 $\pm$ 3.25 & 15,269          \\ \bottomrule
\end{tabular}%
}
\caption{Dataset statistics. Sentences per context are given as mean $\pm$ standard deviation; the total number of sentences equals the number of intervention positions per model and method.}
\label{tab:sent_stats}
\end{table}

\begin{table}[t]
\centering
\small
\begin{tabular}{@{}lp{0.62\columnwidth}@{}}
\toprule
\textbf{Characteristic} & \textbf{Description} \\
\midrule
Length & Number of characters in the sentence. \\
Overlap & Jaccard similarity between the content words of the sentence and of the question (MedQA, MMLU) or claim (DRUID). \\
Readability & Flesch reading ease score. \\
Named entities & Number of named entities in the sentence. \\
Negation & Whether the sentence contains a negation (a syntactic negation or a negation cue such as \textit{no}, \textit{not}, \textit{without}). \\
\midrule
Pronouns & Number of pronouns in the sentence. \\
Pronoun ratio & Share of tokens in the sentence that are pronouns. \\
Coref.\ mentions & Number of mentions in the sentence that belong to a coreference chain in the context. \\
Coref.\ chains & Number of distinct coreference chains with a mention in the sentence. \\
Backward refs. & Number of mentions in the sentence whose chain was first mentioned in an earlier sentence. \\
\bottomrule
\end{tabular}
\caption{Sentence-level lexical characteristics. The first five follow \citet{hagstrom-etal-2025-reality}; the last five capture references across sentences. Coreference is resolved over the whole context with coreferee.}
\label{tab:lexical_characteristics}
\end{table}

\subsection{Implementation of Interventions}\label{appx:implementation}

Table \ref{tab:interv_prompt} presents prompts used to generate intervention texts for \textit{Paraphrase}, \textit{Self-explanation}, and \textit{Teacher-explanation}. Sentences are segmented with spaCy (\texttt{en\_core\_web\_sm}).\footnote{\url{https://spacy.io/models/en\#en_core_web_sm}}
For our experiments, each run (one model, method, and dataset) took one to two hours on an AMD Instinct MI250X GPU. For Qwen-72b, we used an NVIDIA B200 GPU, which took 0.5 hours per run.

\section{Results}\label{appx:results}

Tables \ref{tab:acc_mmlu} and \ref{tab:acc_druid} show sentence-averaged accuracy results on MMLU and DRUID, respectively. On MMLU, context augmentation methods reject the null hypothesis 5 out of 35 times, while DRUID shows the most rejections: 12 out of 35. \textit{Teacher-explanation} shows the most frequent accuracy gains on both datasets. On MMLU, methods that add only computation do not consistently improve performance.

Tables \ref{tab:null_medqa}, \ref{tab:null_mmlu} and \ref{tab:null_druid} show detailed sign-randomisation test results.

\begin{table*}[]
\centering
\resizebox{\textwidth}{!}{%
\begin{tabular}{@{}l@{}}
\toprule
\textbf{Input prompt for \textit{Paraphrase}}                                                                                                                                                                                     \\ \midrule
Paraphrase the following sentence. Only output the paraphrased sentence. \textbackslash{}n\textbackslash{}n Sentence: \%s                                                                                                \\ \midrule
\textbf{Input prompt for \textit{Self-explanation} and \textit{Teacher-explanation}}                                                                                                                                                       \\ \midrule
Generate additional knowledge to understand the sentence marked with \textless{}\textless{} \textgreater{}\textgreater{} in the {[}Context{]}. Only generate a sentence. \\ \textbackslash{}n\textbackslash{}n Context: \%s \\ \bottomrule
\end{tabular}%
}
\caption{We present prompts used for generating intervention texts for \textit{Paraphrase}, \textit{Self-explanation}, and \textit{Teacher-explanation}.}
\label{tab:interv_prompt}
\end{table*}

\begin{table*}[t]
\centering
\resizebox{0.95\textwidth}{!}{%
\begin{tabular}{@{}lrrrrrr@{}}
\toprule
Model & \textbf{\textit{Original}} & \textbf{\textit{Blank}} & \textbf{\textit{Repeat}} & \textbf{\textit{Paraphrase}} & \textbf{\textit{Self-expl.}} & \textbf{\textit{Teacher-expl.}} \\
\midrule
OLMo-1b & 26.8 & \textbf{27.1}\,{\scriptsize $+$2.2/$-$1.8} & \textbf{27.4}\,{\scriptsize $+$2.7/$-$2.1} & \textbf{27.3}\,{\scriptsize $+$3.0/$-$2.4} & \textbf{27.6}\,{\scriptsize $+$4.0/$-$3.1} & 25.9\,{\scriptsize $+$9.8/$-$10.6} \\
Llama-3b & 54.0 & 53.6\,{\scriptsize $+$1.2/$-$1.6} & 53.4\,{\scriptsize $+$2.0/$-$2.6} & \textbf{54.0}\,{\scriptsize $+$3.4/$-$3.3} & 53.3\,{\scriptsize $+$4.0/$-$4.7} & \cellcolor{teal!35}\textbf{55.3}\,{\scriptsize $+$5.7/$-$4.3} \\
Gemma-4b & 52.2 & 51.4\,{\scriptsize $+$1.2/$-$2.0} & 51.7\,{\scriptsize $+$1.7/$-$2.3} & 51.4\,{\scriptsize $+$1.7/$-$2.5} & 51.7\,{\scriptsize $+$3.5/$-$4.0} & \textbf{53.8}\,{\scriptsize $+$6.1/$-$4.5} \\
OLMo-7b & 49.3 & 49.1\,{\scriptsize $+$1.4/$-$1.6} & 49.2\,{\scriptsize $+$2.0/$-$2.1} & \textbf{49.4}\,{\scriptsize $+$2.7/$-$2.6} & \cellcolor{teal!35}\textbf{50.2}\,{\scriptsize $+$5.4/$-$4.5} & \textbf{49.4}\,{\scriptsize $+$9.6/$-$9.5} \\
Llama-8b & 52.6 & \cellcolor{teal!35}\textbf{53.1}\,{\scriptsize $+$1.4/$-$1.0} & 52.2\,{\scriptsize $+$1.6/$-$2.0} & \textbf{52.9}\,{\scriptsize $+$1.9/$-$1.7} & \cellcolor{teal!35}\textbf{53.8}\,{\scriptsize $+$4.3/$-$3.2} & \cellcolor{teal!35}\textbf{54.5}\,{\scriptsize $+$4.8/$-$2.9} \\
Gemma-27b & 67.4 & 67.1\,{\scriptsize $+$1.0/$-$1.3} & 66.9\,{\scriptsize $+$1.1/$-$1.6} & 67.1\,{\scriptsize $+$1.5/$-$1.9} & 67.2\,{\scriptsize $+$3.4/$-$3.7} & \textbf{68.2}\,{\scriptsize $+$4.7/$-$3.9} \\
Qwen-72b & 74.6 & 74.5\,{\scriptsize $+$0.7/$-$0.8} & 74.5\,{\scriptsize $+$1.1/$-$1.2} & 74.3\,{\scriptsize $+$1.2/$-$1.5} & 73.5\,{\scriptsize $+$1.8/$-$3.0} & 74.4\,{\scriptsize $+$2.9/$-$3.1} \\
\bottomrule
\end{tabular}%
}
\caption{Sentence-averaged accuracy (\%) of each method on MMLU. \textit{Original} reports an accuracy without intervention. Small numbers with $+$ are the fraction of intervention positions at which the prediction flips from incorrect to correct ($\mathrm{Corr}$), and those with $-$ are the fraction of the positions at which the prediction flips from correct to incorrect ($\mathrm{Brk}$). We colour cells where the method corrects wrong predictions more often than the 95\% interval of the random distribution, without breaking correct predictions more often than the 95\% interval of the random distribution.}
\label{tab:acc_mmlu}
\end{table*}

\begin{table*}[t]
\centering
\resizebox{0.95\textwidth}{!}{%
\begin{tabular}{@{}lrrrrrr@{}}
\toprule
Model & \textbf{\textit{Original}} & \textbf{\textit{Blank}} & \textbf{\textit{Repeat}} & \textbf{\textit{Paraphrase}} & \textbf{\textit{Self-expl.}} & \textbf{\textit{Teacher-expl.}} \\
\midrule
OLMo-1b & 82.4 & 82.2\,{\scriptsize $+$1.3/$-$1.5} & \cellcolor{teal!35}\textbf{82.5}\,{\scriptsize $+$2.5/$-$2.4} & \textbf{82.5}\,{\scriptsize $+$2.4/$-$2.3} & \cellcolor{teal!35}\textbf{83.1}\,{\scriptsize $+$4.6/$-$4.0} & 80.8\,{\scriptsize $+$2.5/$-$4.2} \\
Llama-3b & 71.1 & \cellcolor{teal!35}\textbf{71.4}\,{\scriptsize $+$1.4/$-$1.1} & \cellcolor{teal!35}\textbf{72.0}\,{\scriptsize $+$2.3/$-$1.4} & \textbf{71.6}\,{\scriptsize $+$2.1/$-$1.7} & 71.1\,{\scriptsize $+$1.8/$-$1.9} & \cellcolor{teal!35}\textbf{72.1}\,{\scriptsize $+$2.6/$-$1.6} \\
Gemma-4b & 87.1 & \cellcolor{teal!35}\textbf{87.1}\,{\scriptsize $+$1.0/$-$1.0} & 86.9\,{\scriptsize $+$1.3/$-$1.5} & 86.6\,{\scriptsize $+$1.2/$-$1.7} & 87.0\,{\scriptsize $+$1.5/$-$1.6} & 86.9\,{\scriptsize $+$1.9/$-$2.1} \\
OLMo-7b & 85.0 & 81.7\,{\scriptsize $+$2.1/$-$5.3} & 82.3\,{\scriptsize $+$2.0/$-$4.7} & 82.0\,{\scriptsize $+$2.1/$-$5.1} & 81.2\,{\scriptsize $+$2.1/$-$6.0} & 77.0\,{\scriptsize $+$2.6/$-$10.6} \\
Llama-8b & 79.4 & 79.2\,{\scriptsize $+$1.1/$-$1.3} & 79.2\,{\scriptsize $+$1.5/$-$1.7} & 79.3\,{\scriptsize $+$1.6/$-$1.7} & \cellcolor{teal!35}\textbf{80.3}\,{\scriptsize $+$2.3/$-$1.4} & \cellcolor{teal!35}\textbf{80.3}\,{\scriptsize $+$2.4/$-$1.5} \\
Gemma-27b & 89.7 & 89.5\,{\scriptsize $+$0.3/$-$0.5} & 89.6\,{\scriptsize $+$0.6/$-$0.7} & 89.4\,{\scriptsize $+$0.5/$-$0.8} & 89.4\,{\scriptsize $+$0.7/$-$1.0} & 89.2\,{\scriptsize $+$0.7/$-$1.2} \\
Qwen-72b & 84.8 & 84.5\,{\scriptsize $+$0.2/$-$0.5} & \cellcolor{teal!35}\textbf{85.2}\,{\scriptsize $+$0.8/$-$0.4} & \cellcolor{teal!35}\textbf{85.1}\,{\scriptsize $+$0.8/$-$0.4} & \cellcolor{teal!35}\textbf{85.5}\,{\scriptsize $+$1.1/$-$0.4} & \cellcolor{teal!35}\textbf{85.3}\,{\scriptsize $+$1.0/$-$0.5} \\
\bottomrule
\end{tabular}%
}
\caption{Sentence-averaged accuracy (\%) of each method on DRUID.}
\label{tab:acc_druid}
\end{table*}

\begin{table*}[]
\centering
\resizebox{\textwidth}{!}{%
\begin{tabular}{@{}l|rr|rr|rr|rr|rr@{}}
\toprule
 & \multicolumn{2}{c|}{\textit{Blank}} & \multicolumn{2}{c|}{\textit{Repeat}} & \multicolumn{2}{c|}{\textit{Paraphrase}} & \multicolumn{2}{c|}{\textit{Self-explanation}} & \multicolumn{2}{c}{\textit{Teacher-explanation}} \\ \midrule
\multicolumn{1}{c|}{} & \multicolumn{1}{c}{Rescue} & \multicolumn{1}{c|}{Break} & \multicolumn{1}{c}{Rescue} & \multicolumn{1}{c|}{Break} & \multicolumn{1}{c}{Rescue} & \multicolumn{1}{c|}{Break} & \multicolumn{1}{c}{Rescue} & \multicolumn{1}{c|}{Break} & \multicolumn{1}{c}{Rescue} & \multicolumn{1}{c}{Break} \\
\textbf{OLMo-1b} & 0.922 & 0.922 & 0.916 & 0.915 & 0.897 & 0.903 & 0.816 & 0.823 & 0.958 & 0.965 \\
\textbf{Llama-3b} & 0.747 & 0.883 & 0.646 & 0.813 & 0.628 & 0.787 & 0.560 & 0.624 & 0.521 & 0.555 \\
\textbf{Gemma-4b} & 0.449 & 0.373 & 0.390 & 0.393 & 0.380 & 0.339 & 0.471 & 0.409 & 0.457 & 0.341 \\
\textbf{OLMo-7b} & 0.549 & 0.521 & 0.475 & 0.464 & 0.419 & 0.463 & 0.421 & 0.412 & 0.461 & 0.451 \\
\textbf{Llama-8b} & 0.761 & 0.895 & 0.701 & 0.844 & 0.679 & 0.825 & 0.550 & 0.636 & 0.504 & 0.637 \\
\textbf{Gemma-27b} & 0.378 & 0.330 & 0.377 & 0.330 & 0.337 & 0.345 & 0.406 & 0.300 & 0.465 & 0.279 \\
\textbf{Qwen-72b} & 0.535 & 0.498 & 0.384 & 0.493 & 0.376 & 0.444 & 0.367 & 0.369 & 0.442 & 0.333 \\ \bottomrule
\end{tabular}%
}
\caption{Multiplicity analysis on MedQA.}
\label{tab:multi_medqa}
\end{table*}

\begin{table*}[]
\centering
\resizebox{\textwidth}{!}{%
\begin{tabular}{@{}l|rr|rr|rr|rr|rr@{}}
\toprule
 & \multicolumn{2}{c|}{\textit{Blank}} & \multicolumn{2}{c|}{\textit{Repeat}} & \multicolumn{2}{c|}{\textit{Paraphrase}} & \multicolumn{2}{c|}{\textit{Self-explanation}} & \multicolumn{2}{c}{\textit{Teacher-explanation}} \\ \midrule
\multicolumn{1}{c|}{} & \multicolumn{1}{c}{Rescue} & \multicolumn{1}{c|}{Break} & \multicolumn{1}{c}{Rescue} & \multicolumn{1}{c|}{Break} & \multicolumn{1}{c}{Rescue} & \multicolumn{1}{c|}{Break} & \multicolumn{1}{c}{Rescue} & \multicolumn{1}{c|}{Break} & \multicolumn{1}{c}{Rescue} & \multicolumn{1}{c}{Break} \\
\textbf{OLMo-1b} & 0.501 & 0.526 & 0.475 & 0.428 & 0.470 & 0.452 & 0.463 & 0.425 & 0.889 & 0.902 \\
\textbf{Llama-3b} & 0.394 & 0.447 & 0.434 & 0.439 & 0.458 & 0.463 & 0.412 & 0.439 & 0.419 & 0.412 \\
\textbf{Gemma-4b} & 0.392 & 0.419 & 0.382 & 0.381 & 0.385 & 0.369 & 0.375 & 0.379 & 0.396 & 0.359 \\
\textbf{OLMo-7b} & 0.336 & 0.436 & 0.417 & 0.427 & 0.397 & 0.403 & 0.414 & 0.420 & 0.474 & 0.489 \\
\textbf{Llama-8b} & 0.430 & 0.395 & 0.433 & 0.418 & 0.400 & 0.377 & 0.433 & 0.384 & 0.374 & 0.348 \\
\textbf{Gemma-27b} & 0.448 & 0.431 & 0.362 & 0.443 & 0.382 & 0.425 & 0.366 & 0.365 & 0.388 & 0.364 \\
\textbf{Qwen-72b} & 0.373 & 0.337 & 0.386 & 0.400 & 0.361 & 0.365 & 0.375 & 0.394 & 0.365 & 0.324 \\ \bottomrule
\end{tabular}%
}
\caption{Multiplicity analysis on MMLU.}
\label{tab:multi_mmlu}
\end{table*}

\begin{table*}[]
\centering
\resizebox{\textwidth}{!}{%
\begin{tabular}{@{}l|rr|rr|rr|rr|rr@{}}
\toprule
 & \multicolumn{2}{c|}{\textit{Blank}} & \multicolumn{2}{c|}{\textit{Repeat}} & \multicolumn{2}{c|}{\textit{Paraphrase}} & \multicolumn{2}{c|}{\textit{Self-explanation}} & \multicolumn{2}{c}{\textit{Teacher-explanation}} \\ \midrule
\multicolumn{1}{c|}{} & \multicolumn{1}{c}{Rescue} & \multicolumn{1}{c|}{Break} & \multicolumn{1}{c}{Rescue} & \multicolumn{1}{c|}{Break} & \multicolumn{1}{c}{Rescue} & \multicolumn{1}{c|}{Break} & \multicolumn{1}{c}{Rescue} & \multicolumn{1}{c|}{Break} & \multicolumn{1}{c}{Rescue} & \multicolumn{1}{c}{Break} \\
\textbf{OLMo-1b} & 0.374 & 0.392 & 0.421 & 0.452 & 0.384 & 0.383 & 0.464 & 0.399 & 0.445 & 0.505 \\
\textbf{Llama-3b} & 0.437 & 0.424 & 0.437 & 0.392 & 0.436 & 0.420 & 0.421 & 0.425 & 0.443 & 0.386 \\
\textbf{Gemma-4b} & 0.377 & 0.336 & 0.441 & 0.390 & 0.396 & 0.372 & 0.413 & 0.349 & 0.456 & 0.358 \\
\textbf{OLMo-7b} & 0.744 & 0.720 & 0.648 & 0.637 & 0.705 & 0.620 & 0.634 & 0.626 & 0.750 & 0.678 \\
\textbf{Llama-8b} & 0.437 & 0.353 & 0.436 & 0.395 & 0.417 & 0.367 & 0.471 & 0.360 & 0.461 & 0.347 \\
\textbf{Gemma-27b} & 0.305 & 0.286 & 0.410 & 0.353 & 0.380 & 0.315 & 0.381 & 0.337 & 0.409 & 0.340 \\
\textbf{Qwen-72b} & 0.456 & 0.392 & 0.401 & 0.344 & 0.393 & 0.349 & 0.421 & 0.284 & 0.443 & 0.297 \\ \bottomrule
\end{tabular}%
}
\caption{Multiplicity analysis on DRUID.}
\label{tab:multi_druid}
\end{table*}

\begin{table*}[t]
\centering
\small
\setlength{\tabcolsep}{4pt}
\begin{tabular}{@{}llrrcrcrrcrc@{}}
\toprule
 & & \multicolumn{4}{c}{$\mathrm{Corr}$} & & \multicolumn{4}{c}{$\mathrm{Brk}$} & \\
\cmidrule(lr){3-6}\cmidrule(lr){8-11}
Method & Model & Obs. & Null & 95\% interval & $\Delta$ & & Obs. & Null & 95\% interval & $\Delta$ & Pass \\
\midrule
Blank & OLMo-1b & 18.8 & 9.0 & [8.5, 9.5] & +9.8 & & 15.3 & 7.3 & [6.9, 7.7] & +8.0 &  \\
 & Llama-3b & 5.4 & 7.8 & [7.4, 8.3] & $-$2.4 & & 10.7 & 6.9 & [6.4, 7.4] & +3.8 &  \\
 & Gemma-4b & 2.1 & 1.7 & [1.5, 1.9] & +0.4 & & 1.1 & 1.1 & [0.9, 1.2] & +0.1 & $\checkmark$ \\
 & OLMo-7b & 3.2 & 2.7 & [2.4, 3.0] & +0.5 & & 3.0 & 2.5 & [2.3, 2.8] & +0.4 &  \\
 & Llama-8b & 5.2 & 6.3 & [5.9, 6.7] & $-$1.2 & & 12.5 & 9.1 & [8.6, 9.6] & +3.4 &  \\
 & Gemma-27b & 1.3 & 1.2 & [1.0, 1.4] & +0.1 & & 0.7 & 0.9 & [0.8, 1.0] & $-$0.2 &  \\
 & Qwen-72b & 1.4 & 1.2 & [1.0, 1.3] & +0.2 & & 1.4 & 1.1 & [0.9, 1.3] & +0.3 &  \\
\midrule
Repeat & OLMo-1b & 19.9 & 9.6 & [9.0, 10.1] & +10.3 & & 16.7 & 8.0 & [7.6, 8.5] & +8.7 &  \\
 & Llama-3b & 5.9 & 8.2 & [7.8, 8.7] & $-$2.4 & & 11.1 & 7.4 & [7.0, 7.9] & +3.7 &  \\
 & Gemma-4b & 2.8 & 2.3 & [2.0, 2.5] & +0.6 & & 1.7 & 1.3 & [1.1, 1.5] & +0.4 &  \\
 & OLMo-7b & 3.5 & 3.0 & [2.7, 3.3] & +0.5 & & 3.2 & 2.6 & [2.3, 2.9] & +0.6 &  \\
 & Llama-8b & 3.9 & 5.9 & [5.5, 6.3] & $-$2.0 & & 14.4 & 9.9 & [9.3, 10.4] & +4.5 &  \\
 & Gemma-27b & 2.0 & 1.6 & [1.4, 1.9] & +0.3 & & 1.3 & 1.2 & [1.1, 1.4] & +0.0 & $\checkmark$ \\
 & Qwen-72b & 1.2 & 1.3 & [1.1, 1.5] & $-$0.1 & & 1.9 & 1.4 & [1.2, 1.6] & +0.5 &  \\
\midrule
Paraphrase & OLMo-1b & 19.9 & 9.6 & [9.1, 10.1] & +10.3 & & 16.7 & 8.1 & [7.7, 8.6] & +8.6 &  \\
 & Llama-3b & 6.2 & 9.1 & [8.7, 9.6] & $-$2.9 & & 11.0 & 8.3 & [7.8, 8.8] & +2.7 &  \\
 & Gemma-4b & 3.0 & 2.4 & [2.1, 2.7] & +0.6 & & 1.8 & 1.4 & [1.2, 1.5] & +0.4 &  \\
 & OLMo-7b & 3.5 & 3.1 & [2.8, 3.4] & +0.4 & & 3.3 & 2.7 & [2.4, 3.0] & +0.6 &  \\
 & Llama-8b & 4.2 & 5.9 & [5.5, 6.3] & $-$1.7 & & 14.2 & 9.7 & [9.2, 10.3] & +4.4 &  \\
 & Gemma-27b & 1.9 & 1.6 & [1.4, 1.8] & +0.3 & & 1.4 & 1.2 & [1.1, 1.4] & +0.1 & $\checkmark$ \\
 & Qwen-72b & 1.4 & 1.4 & [1.2, 1.6] & +0.0 & & 2.0 & 1.5 & [1.3, 1.7] & +0.5 &  \\
\midrule
Self-explanation & OLMo-1b & 18.4 & 9.0 & [8.5, 9.5] & +9.4 & & 14.7 & 7.1 & [6.6, 7.5] & +7.6 &  \\
 & Llama-3b & 10.1 & 10.9 & [10.4, 11.4] & $-$0.8 & & 13.4 & 10.5 & [10.0, 11.0] & +2.9 &  \\
 & Gemma-4b & 8.7 & 5.7 & [5.3, 6.1] & +3.0 & & 6.2 & 4.0 & [3.6, 4.3] & +2.3 &  \\
 & OLMo-7b & 9.4 & 7.9 & [7.4, 8.4] & +1.5 & & 7.0 & 6.0 & [5.6, 6.4] & +1.0 &  \\
 & Llama-8b & 9.4 & 8.2 & [7.8, 8.7] & +1.2 & & 13.7 & 10.9 & [10.3, 11.4] & +2.9 &  \\
 & Gemma-27b & 5.9 & 3.8 & [3.5, 4.2] & +2.0 & & 3.4 & 2.6 & [2.3, 2.9] & +0.8 &  \\
 & Qwen-72b & 3.4 & 2.9 & [2.6, 3.2] & +0.5 & & 3.1 & 3.1 & [2.8, 3.4] & +0.0 & $\checkmark$ \\
\midrule
Teacher-explanation & OLMo-1b & 24.1 & 12.2 & [11.7, 12.8] & +11.8 & & 20.0 & 10.1 & [9.6, 10.7] & +9.9 &  \\
 & Llama-3b & 12.4 & 11.5 & [10.9, 12.0] & +0.9 & & 10.4 & 10.4 & [9.8, 10.9] & +0.0 & $\checkmark$ \\
 & Gemma-4b & 14.1 & 8.2 & [7.8, 8.7] & +5.8 & & 4.7 & 3.5 & [3.2, 3.8] & +1.2 &  \\
 & OLMo-7b & 15.5 & 10.5 & [9.9, 11.0] & +5.1 & & 8.5 & 7.0 & [6.6, 7.5] & +1.5 &  \\
 & Llama-8b & 10.7 & 8.5 & [8.0, 9.0] & +2.2 & & 12.1 & 10.5 & [9.9, 11.0] & +1.7 &  \\
 & Gemma-27b & 9.7 & 5.7 & [5.3, 6.1] & +4.0 & & 2.8 & 2.6 & [2.4, 2.9] & +0.2 & $\checkmark$ \\
 & Qwen-72b & 6.3 & 4.1 & [3.7, 4.5] & +2.2 & & 2.9 & 4.2 & [3.9, 4.6] & $-$1.3 & $\checkmark$ \\
\bottomrule
\end{tabular}
\caption{MedQA: sign-randomisation test for $\mathrm{Corr}$ and $\mathrm{Brk}$ (\%). Obs.: observed value; Null: mean of the null distribution over $B = 1000$ sign-randomised draws, with its 95\% interval; $\Delta$: observed minus null mean. Pass: $\mathrm{Corr}$ exceeds its 95\% interval while $\mathrm{Brk}$ does not.}
\label{tab:null_medqa}
\end{table*}

\begin{table*}[t]
\centering
\small
\setlength{\tabcolsep}{4pt}
\begin{tabular}{@{}llrrcrcrrcrc@{}}
\toprule
 & & \multicolumn{4}{c}{$\mathrm{Corr}$} & & \multicolumn{4}{c}{$\mathrm{Brk}$} & \\
\cmidrule(lr){3-6}\cmidrule(lr){8-11}
Method & Model & Obs. & Null & 95\% interval & $\Delta$ & & Obs. & Null & 95\% interval & $\Delta$ & Pass \\
\midrule
Blank & OLMo-1b & 2.2 & 2.6 & [2.4, 2.8] & $-$0.3 & & 1.8 & 2.0 & [1.8, 2.1] & $-$0.1 &  \\
 & Llama-3b & 1.2 & 1.5 & [1.4, 1.6] & $-$0.3 & & 1.6 & 1.9 & [1.7, 2.0] & $-$0.2 &  \\
 & Gemma-4b & 1.2 & 1.1 & [0.9, 1.2] & +0.1 & & 2.0 & 1.5 & [1.3, 1.7] & +0.5 &  \\
 & OLMo-7b & 1.4 & 1.6 & [1.4, 1.7] & $-$0.1 & & 1.6 & 1.5 & [1.4, 1.7] & +0.1 &  \\
 & Llama-8b & 1.4 & 1.2 & [1.0, 1.3] & +0.2 & & 1.0 & 1.6 & [1.4, 1.8] & $-$0.6 & $\checkmark$ \\
 & Gemma-27b & 1.0 & 0.9 & [0.7, 1.0] & +0.1 & & 1.3 & 1.1 & [0.9, 1.2] & +0.2 &  \\
 & Qwen-72b & 0.7 & 0.8 & [0.7, 0.9] & $-$0.1 & & 0.8 & 0.9 & [0.7, 1.0] & $-$0.1 &  \\
\midrule
Repeat & OLMo-1b & 2.7 & 3.0 & [2.8, 3.2] & $-$0.3 & & 2.1 & 2.4 & [2.2, 2.6] & $-$0.3 &  \\
 & Llama-3b & 2.0 & 1.9 & [1.8, 2.1] & +0.0 & & 2.6 & 2.6 & [2.4, 2.8] & +0.0 &  \\
 & Gemma-4b & 1.7 & 1.3 & [1.1, 1.5] & +0.4 & & 2.3 & 1.7 & [1.5, 1.9] & +0.5 &  \\
 & OLMo-7b & 2.0 & 2.0 & [1.8, 2.2] & +0.0 & & 2.1 & 1.9 & [1.7, 2.2] & +0.1 &  \\
 & Llama-8b & 1.6 & 1.8 & [1.6, 1.9] & $-$0.1 & & 2.0 & 2.1 & [1.9, 2.3] & +0.0 &  \\
 & Gemma-27b & 1.1 & 1.1 & [0.9, 1.2] & +0.0 & & 1.6 & 1.4 & [1.2, 1.6] & +0.2 &  \\
 & Qwen-72b & 1.1 & 1.2 & [1.0, 1.3] & $-$0.1 & & 1.2 & 1.3 & [1.2, 1.5] & $-$0.1 &  \\
\midrule
Paraphrase & OLMo-1b & 3.0 & 3.3 & [3.0, 3.5] & $-$0.3 & & 2.4 & 2.5 & [2.3, 2.7] & $-$0.1 &  \\
 & Llama-3b & 3.4 & 3.1 & [2.9, 3.4] & +0.2 & & 3.3 & 3.7 & [3.5, 4.0] & $-$0.4 &  \\
 & Gemma-4b & 1.7 & 1.4 & [1.3, 1.6] & +0.2 & & 2.5 & 1.9 & [1.7, 2.0] & +0.6 &  \\
 & OLMo-7b & 2.7 & 2.6 & [2.4, 2.9] & +0.1 & & 2.6 & 2.4 & [2.2, 2.6] & +0.2 &  \\
 & Llama-8b & 1.9 & 1.8 & [1.6, 2.0] & +0.1 & & 1.7 & 2.2 & [2.0, 2.4] & $-$0.5 &  \\
 & Gemma-27b & 1.5 & 1.2 & [1.1, 1.4] & +0.3 & & 1.9 & 1.5 & [1.3, 1.7] & +0.4 &  \\
 & Qwen-72b & 1.2 & 1.3 & [1.2, 1.5] & $-$0.2 & & 1.5 & 1.6 & [1.4, 1.7] & $-$0.1 &  \\
\midrule
Self-explanation & OLMo-1b & 4.0 & 4.0 & [3.7, 4.3] & $-$0.1 & & 3.1 & 3.2 & [3.0, 3.4] & $-$0.1 &  \\
 & Llama-3b & 4.0 & 4.1 & [3.8, 4.4] & $-$0.1 & & 4.7 & 5.0 & [4.7, 5.3] & $-$0.3 &  \\
 & Gemma-4b & 3.5 & 2.4 & [2.2, 2.6] & +1.0 & & 4.0 & 2.8 & [2.5, 3.0] & +1.2 &  \\
 & OLMo-7b & 5.4 & 5.0 & [4.6, 5.3] & +0.5 & & 4.5 & 4.3 & [4.0, 4.6] & +0.2 & $\checkmark$ \\
 & Llama-8b & 4.3 & 3.7 & [3.5, 4.0] & +0.5 & & 3.2 & 4.0 & [3.7, 4.2] & $-$0.8 & $\checkmark$ \\
 & Gemma-27b & 3.4 & 2.5 & [2.3, 2.7] & +0.9 & & 3.7 & 2.6 & [2.3, 2.8] & +1.1 &  \\
 & Qwen-72b & 1.8 & 2.2 & [2.0, 2.4] & $-$0.4 & & 3.0 & 2.8 & [2.5, 3.0] & +0.2 &  \\
\midrule
Teacher-explanation & OLMo-1b & 9.8 & 12.3 & [11.8, 12.8] & $-$2.5 & & 10.6 & 9.9 & [9.4, 10.3] & +0.7 &  \\
 & Llama-3b & 5.7 & 4.8 & [4.5, 5.1] & +0.9 & & 4.3 & 5.8 & [5.5, 6.2] & $-$1.5 & $\checkmark$ \\
 & Gemma-4b & 6.1 & 3.8 & [3.5, 4.0] & +2.3 & & 4.5 & 3.3 & [3.0, 3.5] & +1.2 &  \\
 & OLMo-7b & 9.6 & 8.0 & [7.6, 8.3] & +1.7 & & 9.5 & 6.9 & [6.5, 7.3] & +2.6 &  \\
 & Llama-8b & 4.8 & 4.0 & [3.7, 4.2] & +0.8 & & 2.9 & 4.1 & [3.8, 4.4] & $-$1.2 & $\checkmark$ \\
 & Gemma-27b & 4.7 & 3.0 & [2.8, 3.3] & +1.6 & & 3.9 & 2.8 & [2.6, 3.1] & +1.0 &  \\
 & Qwen-72b & 2.9 & 2.6 & [2.4, 2.9] & +0.2 & & 3.1 & 3.7 & [3.5, 4.0] & $-$0.7 &  \\
\bottomrule
\end{tabular}
\caption{MMLU: sign-randomisation test for $\mathrm{Corr}$ and $\mathrm{Brk}$ (\%). Obs.: observed value; Null: mean of the null distribution over $B = 1000$ sign-randomised draws, with its 95\% interval; $\Delta$: observed minus null mean. Pass: $\mathrm{Corr}$ exceeds its 95\% interval while $\mathrm{Brk}$ does not.}
\label{tab:null_mmlu}
\end{table*}

\begin{table*}[t]
\centering
\small
\setlength{\tabcolsep}{4pt}
\begin{tabular}{@{}llrrcrcrrcrc@{}}
\toprule
 & & \multicolumn{4}{c}{$\mathrm{Corr}$} & & \multicolumn{4}{c}{$\mathrm{Brk}$} & \\
\cmidrule(lr){3-6}\cmidrule(lr){8-11}
Method & Model & Obs. & Null & 95\% interval & $\Delta$ & & Obs. & Null & 95\% interval & $\Delta$ & Pass \\
\midrule
Blank & OLMo-1b & 1.3 & 1.6 & [1.4, 1.7] & $-$0.3 & & 1.5 & 1.3 & [1.2, 1.5] & +0.2 &  \\
 & Llama-3b & 1.4 & 1.1 & [1.0, 1.3] & +0.3 & & 1.1 & 1.3 & [1.1, 1.4] & $-$0.1 & $\checkmark$ \\
 & Gemma-4b & 1.0 & 0.8 & [0.6, 0.9] & +0.2 & & 1.0 & 1.1 & [0.9, 1.2] & $-$0.1 & $\checkmark$ \\
 & OLMo-7b & 2.1 & 2.7 & [2.5, 2.9] & $-$0.7 & & 5.3 & 4.4 & [4.1, 4.6] & +1.0 &  \\
 & Llama-8b & 1.1 & 1.1 & [0.9, 1.2] & +0.0 & & 1.3 & 1.5 & [1.3, 1.7] & $-$0.2 &  \\
 & Gemma-27b & 0.3 & 0.3 & [0.2, 0.4] & +0.0 & & 0.5 & 0.5 & [0.4, 0.5] & +0.0 &  \\
 & Qwen-72b & 0.2 & 0.3 & [0.2, 0.3] & +0.0 & & 0.5 & 0.3 & [0.2, 0.4] & +0.1 &  \\
\midrule
Repeat & OLMo-1b & 2.5 & 2.1 & [1.9, 2.3] & +0.4 & & 2.4 & 2.3 & [2.1, 2.5] & +0.1 & $\checkmark$ \\
 & Llama-3b & 2.3 & 1.8 & [1.6, 2.0] & +0.5 & & 1.4 & 1.9 & [1.7, 2.1] & $-$0.5 & $\checkmark$ \\
 & Gemma-4b & 1.3 & 1.1 & [0.9, 1.2] & +0.3 & & 1.5 & 1.3 & [1.1, 1.5] & +0.2 &  \\
 & OLMo-7b & 2.0 & 2.5 & [2.3, 2.7] & $-$0.5 & & 4.7 & 4.0 & [3.7, 4.2] & +0.8 &  \\
 & Llama-8b & 1.5 & 1.4 & [1.2, 1.6] & +0.1 & & 1.7 & 2.1 & [1.9, 2.3] & $-$0.4 &  \\
 & Gemma-27b & 0.6 & 0.5 & [0.4, 0.6] & +0.1 & & 0.7 & 0.7 & [0.6, 0.8] & +0.0 &  \\
 & Qwen-72b & 0.8 & 0.5 & [0.5, 0.6] & +0.3 & & 0.4 & 0.6 & [0.4, 0.7] & $-$0.2 & $\checkmark$ \\
\midrule
Paraphrase & OLMo-1b & 2.4 & 2.2 & [2.0, 2.4] & +0.2 & & 2.3 & 2.1 & [1.9, 2.3] & +0.2 &  \\
 & Llama-3b & 2.1 & 2.1 & [1.9, 2.2] & +0.1 & & 1.7 & 2.7 & [2.4, 2.9] & $-$1.0 &  \\
 & Gemma-4b & 1.2 & 1.0 & [0.8, 1.1] & +0.2 & & 1.7 & 1.4 & [1.2, 1.6] & +0.3 &  \\
 & OLMo-7b & 2.1 & 2.6 & [2.4, 2.8] & $-$0.5 & & 5.1 & 4.3 & [4.1, 4.6] & +0.8 &  \\
 & Llama-8b & 1.6 & 1.4 & [1.3, 1.6] & +0.1 & & 1.7 & 1.9 & [1.7, 2.1] & $-$0.2 &  \\
 & Gemma-27b & 0.5 & 0.5 & [0.4, 0.6] & +0.0 & & 0.8 & 0.7 & [0.6, 0.8] & +0.1 &  \\
 & Qwen-72b & 0.8 & 0.6 & [0.5, 0.7] & +0.2 & & 0.4 & 0.5 & [0.4, 0.7] & $-$0.1 & $\checkmark$ \\
\midrule
Self-explanation & OLMo-1b & 4.6 & 3.5 & [3.3, 3.8] & +1.1 & & 4.0 & 3.9 & [3.7, 4.2] & +0.1 & $\checkmark$ \\
 & Llama-3b & 1.8 & 1.8 & [1.6, 1.9] & +0.1 & & 1.9 & 2.2 & [2.0, 2.4] & $-$0.4 &  \\
 & Gemma-4b & 1.5 & 1.1 & [1.0, 1.3] & +0.4 & & 1.6 & 1.4 & [1.3, 1.6] & +0.2 &  \\
 & OLMo-7b & 2.1 & 2.7 & [2.5, 2.9] & $-$0.5 & & 6.0 & 5.3 & [4.9, 5.5] & +0.7 &  \\
 & Llama-8b & 2.3 & 1.8 & [1.6, 2.0] & +0.5 & & 1.4 & 2.4 & [2.2, 2.6] & $-$1.0 & $\checkmark$ \\
 & Gemma-27b & 0.7 & 0.6 & [0.5, 0.7] & +0.1 & & 1.0 & 0.8 & [0.7, 0.9] & +0.1 &  \\
 & Qwen-72b & 1.1 & 0.7 & [0.6, 0.8] & +0.4 & & 0.4 & 0.6 & [0.5, 0.7] & $-$0.2 & $\checkmark$ \\
\midrule
Teacher-explanation & OLMo-1b & 2.5 & 3.2 & [2.9, 3.4] & $-$0.6 & & 4.2 & 3.5 & [3.2, 3.7] & +0.7 &  \\
 & Llama-3b & 2.6 & 2.1 & [1.9, 2.2] & +0.5 & & 1.6 & 2.4 & [2.2, 2.6] & $-$0.8 & $\checkmark$ \\
 & Gemma-4b & 1.9 & 1.3 & [1.1, 1.4] & +0.6 & & 2.1 & 1.7 & [1.5, 1.9] & +0.3 &  \\
 & OLMo-7b & 2.6 & 3.5 & [3.3, 3.7] & $-$0.9 & & 10.6 & 9.8 & [9.4, 10.2] & +0.8 &  \\
 & Llama-8b & 2.4 & 1.8 & [1.6, 2.0] & +0.6 & & 1.5 & 2.5 & [2.3, 2.8] & $-$1.0 & $\checkmark$ \\
 & Gemma-27b & 0.7 & 0.7 & [0.6, 0.8] & +0.0 & & 1.2 & 1.0 & [0.9, 1.2] & +0.2 &  \\
 & Qwen-72b & 1.0 & 0.7 & [0.6, 0.8] & +0.3 & & 0.5 & 0.6 & [0.5, 0.7] & $-$0.2 & $\checkmark$ \\
\bottomrule
\end{tabular}
\caption{DRUID: sign-randomisation test for $\mathrm{Corr}$ and $\mathrm{Brk}$ (\%). Obs.: observed value; Null: mean of the null distribution over $B = 1000$ sign-randomised draws, with its 95\% interval; $\Delta$: observed minus null mean. Pass: $\mathrm{Corr}$ exceeds its 95\% interval while $\mathrm{Brk}$ does not.}
\label{tab:null_druid}
\end{table*}

\section{Analysis}

\subsection{Margin Analysis Results}\label{appx:margin}

\begin{figure*}
    \centering
         \includegraphics[width=\textwidth]{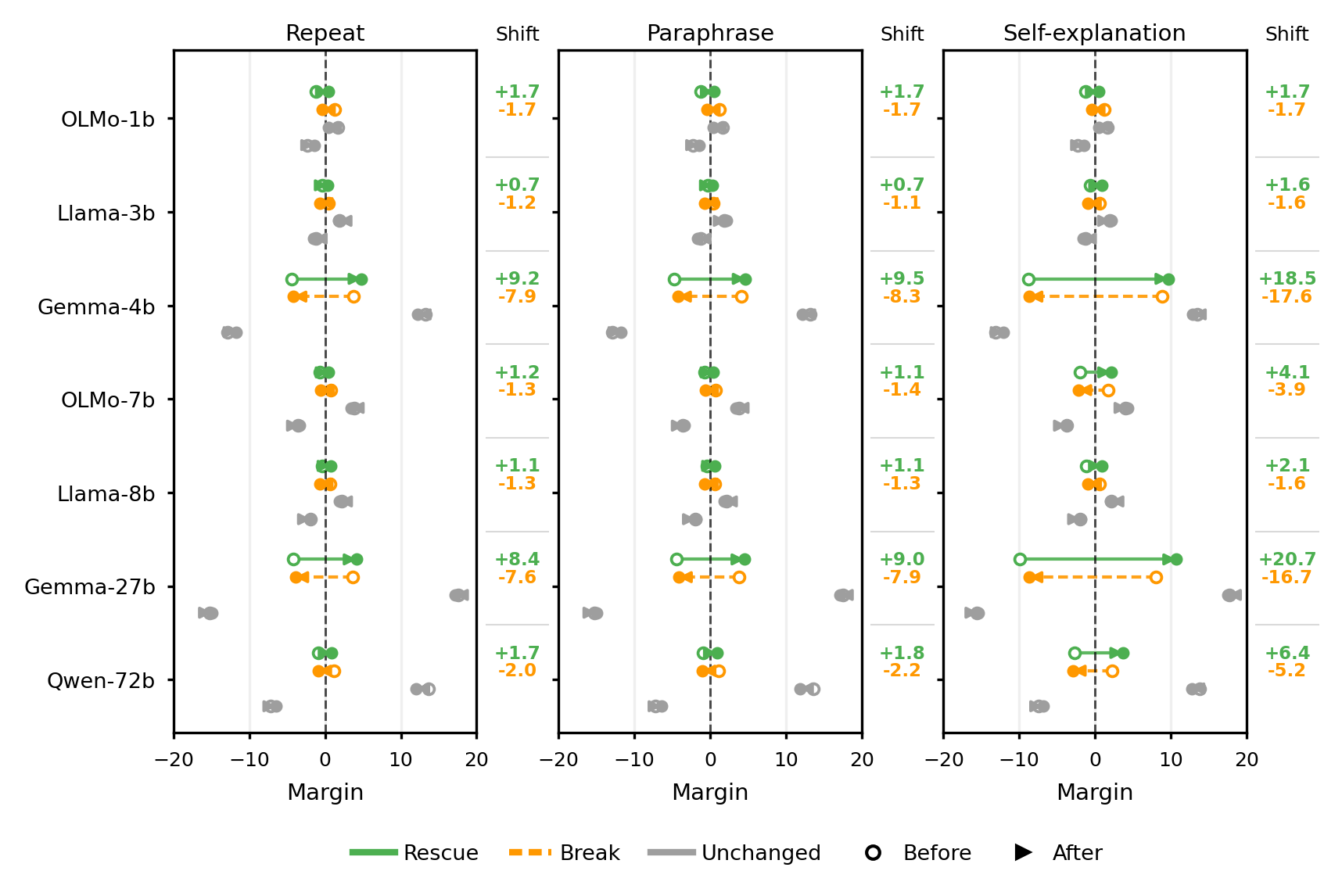}
         \caption{Margin analysis results on MedQA with \textit{Repeat}, \textit{Paraphrase}, and \textit{Self-explanation}.}
         \label{fig:medqa_margin_2}
\end{figure*}

\begin{figure*}
    \centering
         \includegraphics[width=\textwidth]{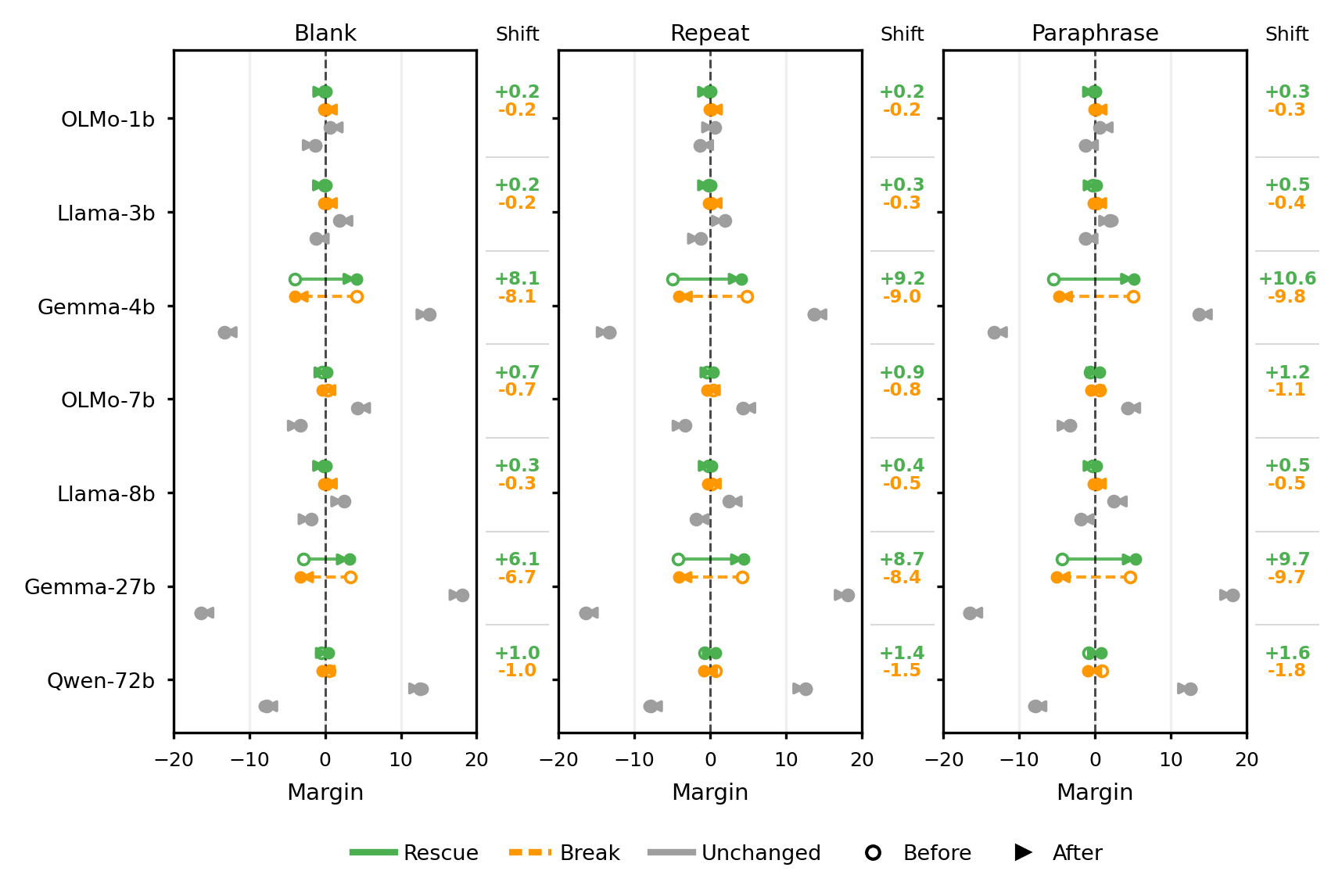}
         \caption{Margin analysis results on MMLU with \textit{Blank}, \textit{Repeat}, and \textit{Paraphrase}.}
         \label{fig:mmlu_margin_1}
\end{figure*}

\begin{figure*}
    \centering
         \includegraphics[width=\textwidth]{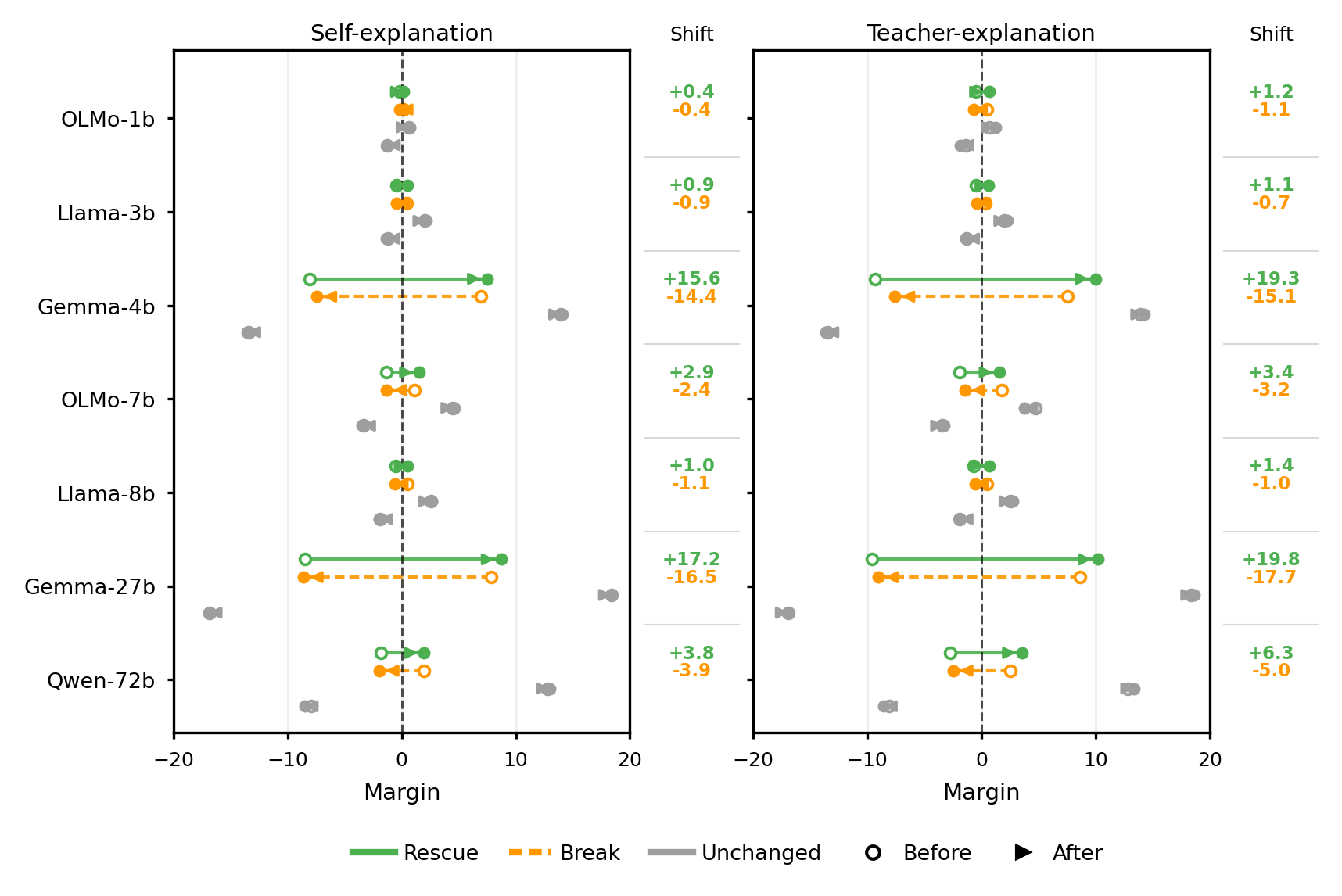}
         \caption{Margin analysis results on MMLU with \textit{Self-explanation} and \textit{Teacher-explanation}.}
         \label{fig:mmlu_margin_2}
\end{figure*}

\begin{figure*}
    \centering
         \includegraphics[width=\textwidth]{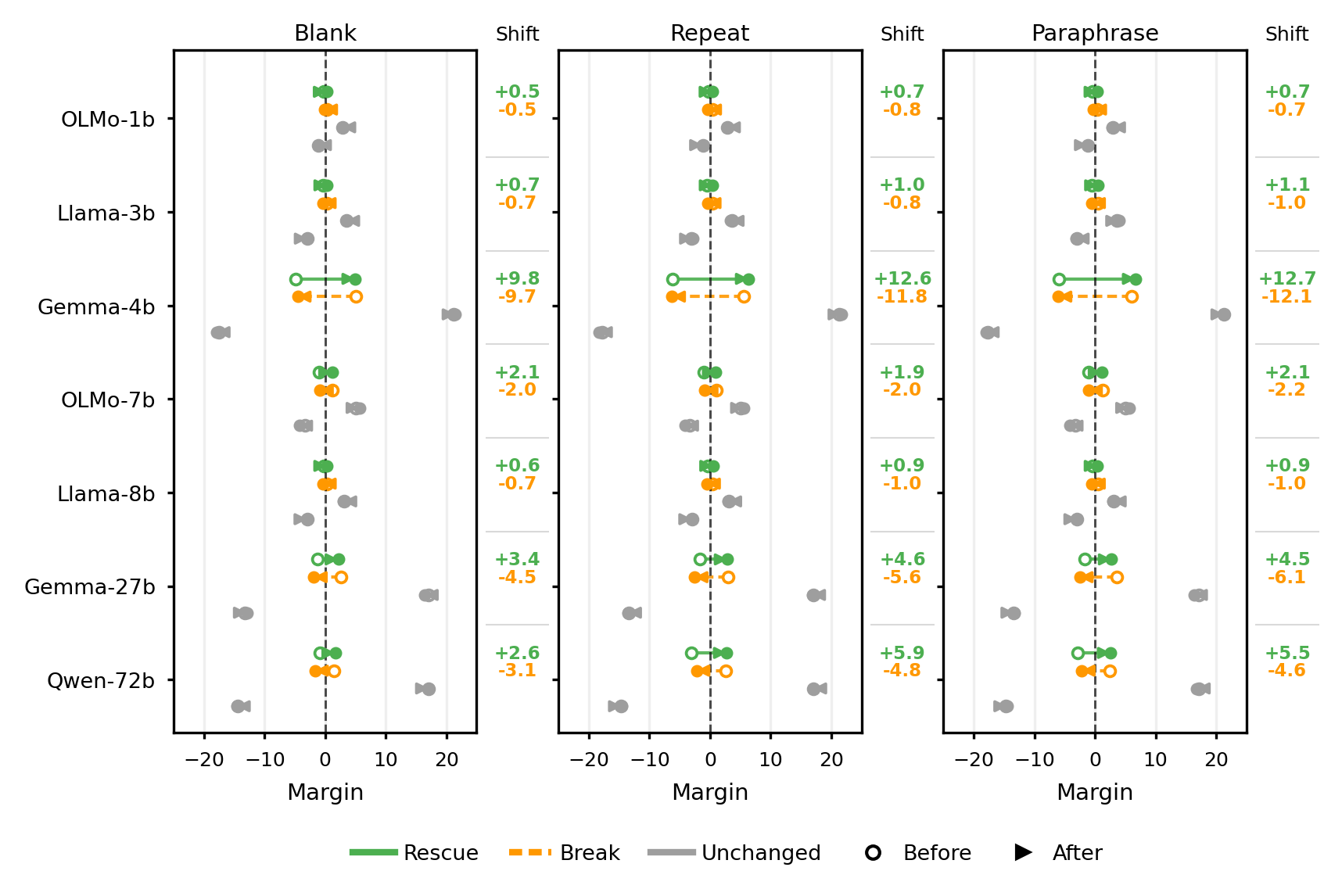}
         \caption{Margin analysis results on DRUID with \textit{Blank}, \textit{Repeat}, and \textit{Paraphrase}.}
         \label{fig:druid_margin_1}
\end{figure*}

\begin{figure*}
    \centering
         \includegraphics[width=\textwidth]{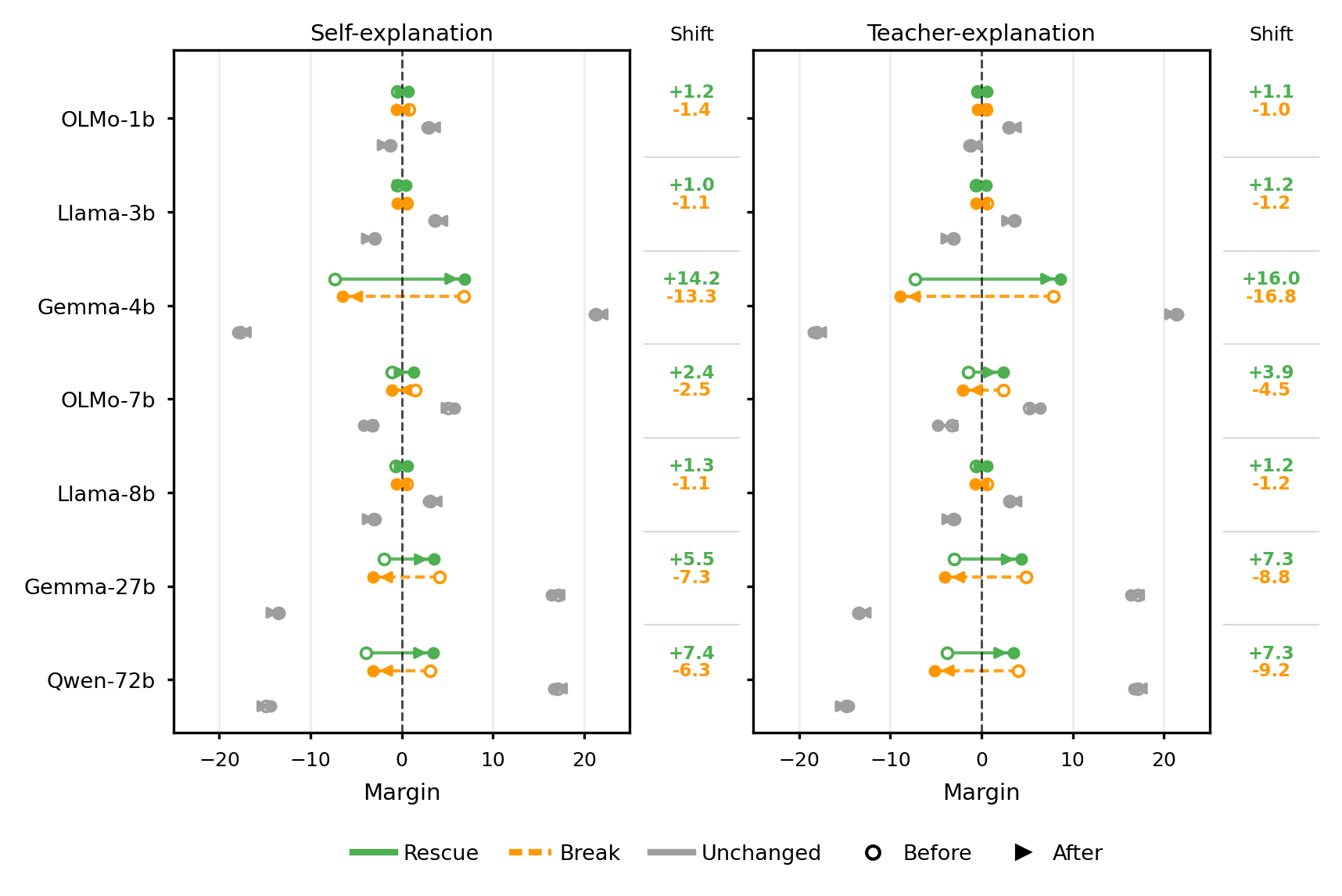}
         \caption{Margin analysis results on DRUID with \textit{Self-explanation} and \textit{Teacher-explanation}.}
         \label{fig:druid_margin_2}
\end{figure*}

Figures \ref{fig:medqa_margin_2}, \ref{fig:mmlu_margin_1}, \ref{fig:mmlu_margin_2}, \ref{fig:druid_margin_1}, and \ref{fig:druid_margin_2} show margin analysis results that are not presented in the main paper. Methods that add both computation and new content (\textit{Self-explanation} and \textit{Teacher-explanation}) yield bigger changes in margins than the methods that add only computation.

\subsection{Multiplicity Analysis}\label{appx:multi}

Tables \ref{tab:multi_medqa}, \ref{tab:multi_mmlu} and \ref{tab:multi_druid} show detailed multiplicity results. We observe similar \texttt{Rescue} and \texttt{Break} rates across models and methods.

\subsection{Predictability with Lexical Characteristics}\label{appx:lexical}

We compute all characteristics on exactly the sentences after which the interventions are inserted, so that each value corresponds to one intervention position.

\paragraph{Computation of Characteristics.}
\textit{Length} is the number of characters in the sentence.
\textit{Overlap} is the Jaccard similarity between the sets of lowercased word tokens of the sentence and of the question (MedQA, MMLU) or claim (DRUID), after removing stopwords.
\textit{Readability} is the Flesch reading ease score, computed with the \texttt{textstat} library.\footnote{\url{https://github.com/textstat/textstat}}
\textit{Named entities} is the number of entities recognised by spaCy (\texttt{en\_core\_web\_lg}).\footnote{\url{https://spacy.io/models/en\#en_core_web_lg}}
\textit{Negation} indicates whether the sentence contains a syntactic negation (a \texttt{neg} dependency) or a lexical negation cue (e.g., \textit{no}, \textit{not}, \textit{without}, \textit{denies}).
\textit{Pronouns} is the number of pronoun tokens in the sentence, and \textit{pronoun ratio} their share among all tokens.
For the coreference characteristics, we run coreferee over the full context and assign each mention to the sentence containing it. \textit{Coref.\ mentions} is the number of mentions in the sentence that belong to a coreference chain, \textit{coref.\ chains} the number of distinct chains with a mention in the sentence, and \textit{backward refs.} the number of mentions whose chain was first mentioned in an earlier sentence.

\paragraph{AUROC.}
For each characteristic, model, method, and dataset, we label each intervention position by whether the intervention leads to a \texttt{Rescue} (or a \texttt{Break}) and compute the AUROC of the characteristic for separating these positions from all others, pooled across contexts (ties count as 0.5). 95\% confidence intervals are obtained by bootstrapping over contexts (2,000 resamples), keeping all positions of a context together.

Table \ref{tab:lexical_auroc} shows AUROC results, where most values are near 0.5, indicating no relation.

\begin{table*}[t]
\centering
\footnotesize
\setlength{\tabcolsep}{3pt}
\begin{tabular}{@{}lcccccccccc@{}}
\toprule
 & \multicolumn{2}{c}{\textit{Blank}} & \multicolumn{2}{c}{\textit{Repeat}} & \multicolumn{2}{c}{\textit{Paraphrase}} & \multicolumn{2}{c}{\textit{Self-explanation}} & \multicolumn{2}{c}{\textit{Teacher-explanation}} \\
\cmidrule(lr){2-3}\cmidrule(lr){4-5}\cmidrule(lr){6-7}\cmidrule(lr){8-9}\cmidrule(lr){10-11}
\textbf{Characteristic} & R & B & R & B & R & B & R & B & R & B \\
\midrule
Length & 0.51\, & 0.51\, & 0.54\, & 0.52\, & 0.53\, & 0.52\, & 0.50\, & 0.49\, & 0.51\, & 0.49\, \\
Overlap & 0.51\, & 0.51\, & 0.52\, & 0.52\, & 0.52\, & 0.52\, & 0.50\, & 0.50\, & 0.51\, & 0.51\, \\
Readability & 0.49\, & 0.50\, & 0.49\, & 0.50\, & 0.48\, & 0.50\, & 0.50\, & 0.51\, & 0.50\, & 0.51\, \\
Named entities & 0.51\, & 0.50\, & 0.53\, & 0.52\, & 0.53\, & 0.52\, & 0.51\, & 0.50\, & 0.51\, & 0.50\, \\
Negation & 0.48\, & 0.50\, & 0.49\, & 0.50\, & 0.49\, & 0.49\, & 0.48\, & 0.49\, & 0.48\, & 0.49\, \\
Pronouns & 0.49\, & 0.50\, & 0.50\, & 0.49\, & 0.50\, & 0.49\, & 0.49\, & 0.48\, & 0.49\, & 0.49\, \\
Pronoun ratio & 0.49\, & 0.49\, & 0.49\, & 0.49\, & 0.49\, & 0.49\, & 0.49\, & 0.48\, & 0.48\, & 0.49\, \\
Coref.\ mentions & 0.51\, & 0.52\, & 0.52\, & 0.52\, & 0.51\, & 0.52\, & 0.50\, & 0.51\, & 0.51\, & 0.51\, \\
Coref.\ chains & 0.51\, & 0.52\, & 0.52\, & 0.52\, & 0.51\, & 0.52\, & 0.50\, & 0.51\, & 0.51\, & 0.51\, \\
Backward refs. & 0.51\, & 0.51\, & 0.51\, & 0.52\, & 0.51\, & 0.52\, & 0.50\, & 0.51\, & 0.50\, & 0.51\, \\
\bottomrule
\end{tabular}
\caption{AUROC (mean over seven models and three datasets) of each lexical characteristic for identifying intervention positions that lead to a \texttt{Rescue} (R) or a \texttt{Break} (B). AUROCs are computed over all intervention positions pooled across contexts; 0.5 indicates no relation.}
\label{tab:lexical_auroc}
\end{table*}

\end{document}